\documentclass[runningheads]{llncs}
\usepackage{eccv}

\usepackage{eccvabbrv}

\usepackage{graphicx}
\usepackage{booktabs}

\usepackage{adjustbox}
\usepackage{rotating}

\usepackage{array}
\usepackage{multirow}
\usepackage[labelformat=parens]{subcaption}

\usepackage{float}

\usepackage[accsupp]{axessibility}  

\usepackage{hyperref}

\usepackage{orcidlink}

\begin{document}


\title{Accelerating Image Super-Resolution \\
Networks with Pixel-Level Classification} 
\titlerunning{Accelerating Image SR \\
Networks with Pixel-Level Classification}

\author{Jinho Jeong\inst{1}\orcidlink{0009-0004-0947-0508} \and
Jinwoo Kim\inst{1}\orcidlink{0009-0001-3250-1788} \and
Younghyun Jo\inst{2}\orcidlink{0000-0002-8530-9802} \and
Seon Joo Kim\inst{1}\orcidlink{0000-0001-8512-216X}}

\authorrunning{J. Jeong et al.}

\institute{Yonsei University \and Samsung Advanced Institute of Technology}

\maketitle
\begin{abstract}
In recent times, the need for effective super-resolution (SR) techniques has surged, especially for large-scale images ranging 2K to 8K resolutions. For DNN-based SISR, decomposing images into overlapping patches is typically necessary due to computational constraints. In such patch-decomposing scheme, one can allocate computational resources differently based on each patch’s difficulty to further improve efficiency while maintaining SR performance. However, this approach has a limitation: computational resources is uniformly allocated within a patch, leading to lower efficiency when the patch contain pixels with varying levels of restoration difficulty. To address the issue, we propose the Pixel-level Classifier for Single Image Super-Resolution (PCSR), a novel method designed to distribute computational resources adaptively at the pixel level. A PCSR model comprises a backbone, a pixel-level classifier, and a set of pixel-level upsamplers with varying capacities. The pixel-level classifier assigns each pixel to an appropriate upsampler based on its restoration difficulty, thereby optimizing computational resource usage. Our method allows for performance and computational cost balance during inference without re-training. Our experiments demonstrate PCSR’s advantage over existing patch-distributing methods in PSNR-FLOP trade-offs across different backbone models and benchmarks. The code is available at
\href{https://github.com/3587jjh/PCSR}{https://github.com/3587jjh/PCSR}.
\end{abstract}
\section{Introduction}
\label{sec:intro}

Single Image Super-Resolution (SISR) is a task focused on restoring a high-resolution (HR) image from its low-resolution (LR) counterpart. 
The task has wide real-life applications across diverse fields, including but not limited to digital photography, medical imaging, surveillance, and security. 
In line with these significant demands, SISR has advanced in last decades, especially with Deep Neural Networks (DNNs) \cite{dong2015image, kim2016accurate, ledig2017photo,  lim2017enhanced, zhang2018image, zhang2018residual}. 

However, as the new SISR models come out, both capacity and computational cost tend to go up, making it hard to apply the models in real-world applications or devices with limited resources. 
Therefore, it has led to a shift towards designing simpler, efficient lightweight models \cite{dong2016accelerating, tai2017memnet, ahn2018fast, zhao2020efficient, gao2022lightweight, li2022blueprint} that consider a balance between performance and computational cost.
In addition, extensive researches \cite{liu2020deep, xie2021learning, hu2022restore, kong2021classsr, chen2022arm, wang2022adaptive} have been developed to reduce the parameter size and/or the number of floating-point operations (FLOPs) of existing models without compromising their performance.

In parallel, there has been a growing demand for efficient SR, particularly with the rise of platforms that provide large-scale images for users such as advanced smartphones, high-definition televisions, or professional-grade monitors that support resolutions ranging from 2K to 8K. 
Nevertheless, SR on a large image is challenging; a large image cannot be processed in a single pass (\ie, \textit{per-image processing}) due to the limitation in computational resources.
Instead, a common approach for large image SR involves dividing a given LR image into overlapping patches, applying an SR model to each patch independently, and then merging the outputs to obtain a super-resolved image.
Several studies \cite{kong2021classsr, chen2022arm, wang2022adaptive} have explored the approach, namely \textit{per-patch processing} approach, with the aim of enhancing the efficiency of existing models while preserving their performance. 
These studies share the observations that each patch varies in restoration difficulty, thus allocating different computational resources to each patch.

\begin{figure*}[t]
    \centering
    \includegraphics[width=\linewidth]{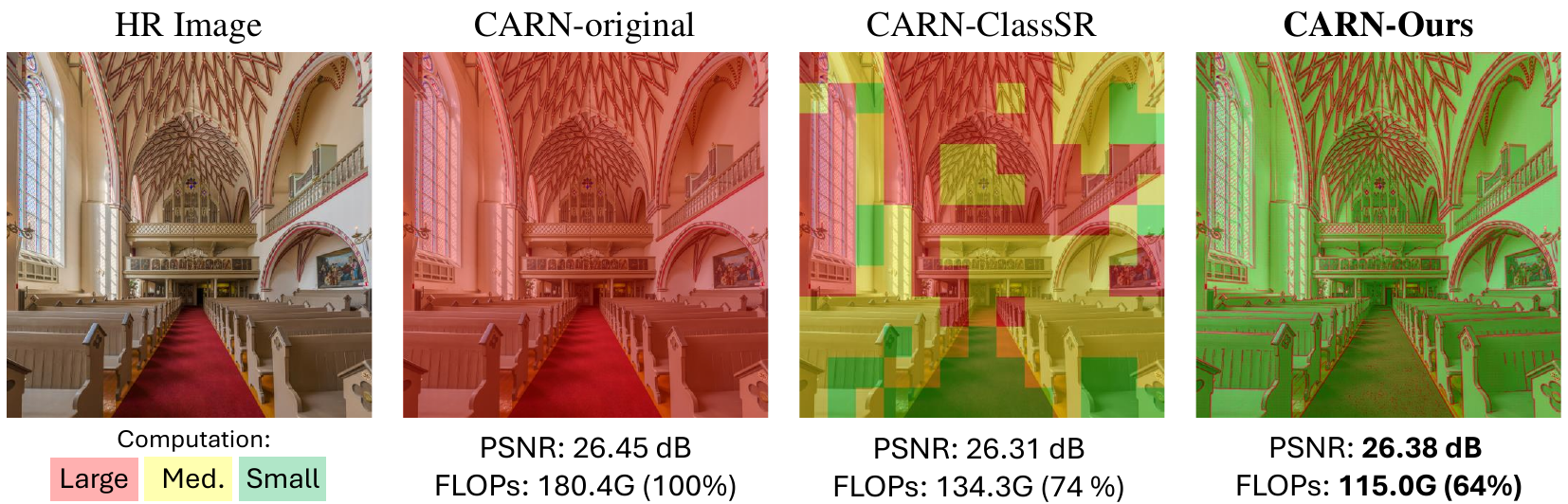}
    \caption{
    The SR result on the image ``1228'' (Test2K), $\times$4.
    By adaptively distributing computational resources in a pixel-wise manner, our method can reduce the overall computational costs in terms of FLOPs compared to the patch-distributing method, while also achieving a better PSNR score.
    }
    \label{fig:teaser1}
\end{figure*}

While adaptively distributing computational resources at the patch-level achieves remarkable improvements of efficiency, it has two limitations that may prevent it from fully leveraging the potential for higher efficiency:
1) Since SR is a low-level vision task, even a single patch can contain pixels with varying degrees of restoration difficulty. 
That is, when allocating large computational resources to a patch that includes easy pixels, it can lead to a waste of computational effort. 
Conversely, if a patch with a smaller allocation of computational resources contains hard pixels, it would negatively impact performance. 
2) These so-called \textit{patch-distributing} methods become less efficient with larger patch sizes, as they are more likely to contain a balanced mix of easy and hard pixels. 
It introduces a dilemma: we may want to use larger patches since it not only minimizes redundant operations from overlapping but also enhances performance by leveraging more contextual information.

In this paper, our primary goal is to enhance the efficiency of existing SISR models, especially for larger images. 
To overcome the aforementioned limitations from patch-distributing methods, we propose a novel approach named Pixel-level Classifier for Single Image Super-Resolution (PCSR), which is specifically designed to adaptively distribute computational resources at the pixel-level.
The model based on our method consists of three main parts: a backbone, a pixel-level classifier, and a set of pixel-level upsamplers with varying capacity. The model operates as follows:
1) The backbone takes an LR input and generates an LR feature map.
2) For each pixel in the HR space, the pixel-level classifier predicts the probability of assigning it to the specific upsampler using the LR feature map and the relative position of that pixel.
3) Accordingly, each pixel is assigned adaptively to a properly sized pixel-level upsampler to predict its RGB value.
4) Finally, super-resolved output is obtained by aggregating the RGB values of every pixels.

\begin{figure*}[t]
    \centering
    \includegraphics[width=\linewidth]{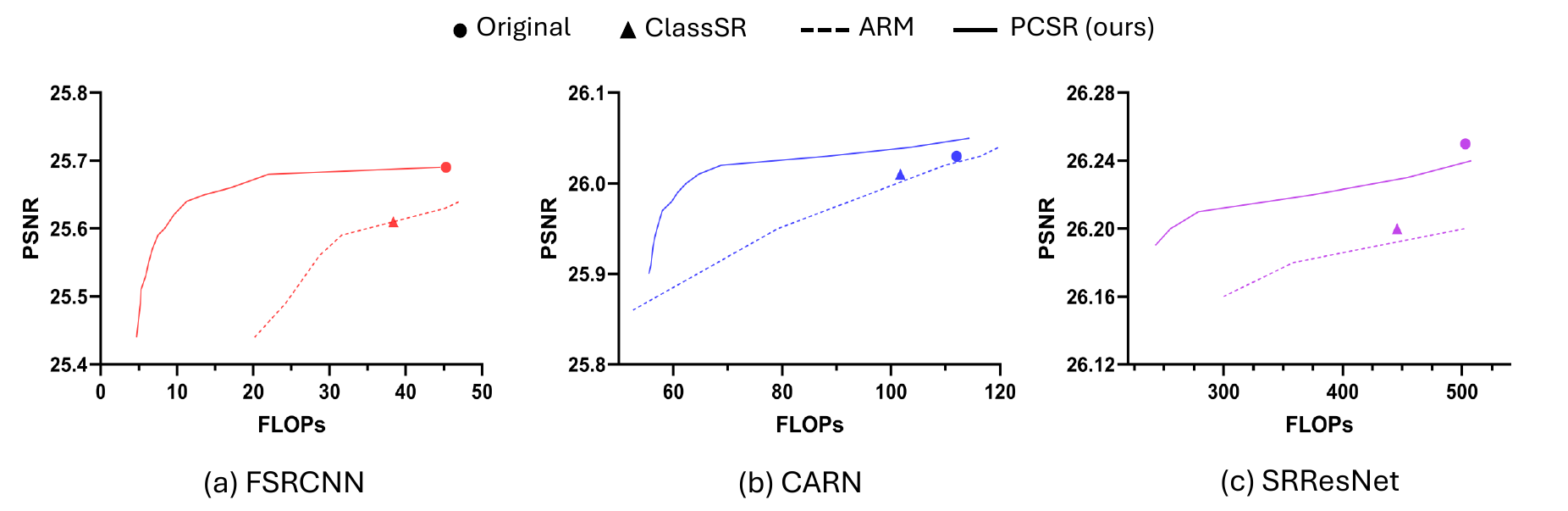}
    \caption{
    Visual comparison of PSNR and FLOPs between ClassSR, ARM, and PCSR (ours) on Test2K at scale $\times$4.
    }
    \label{fig:teaser2}
\end{figure*}

To the best of our knowledge, our method is the first to apply a pixel-wise distributing method in the context of efficient SR for large images.
By cutting down redundant computations in a pixel-wise manner, we can further improve the efficiency of the patch-distributing approach, as illustrated in Fig. \ref{fig:teaser1}. During the inference phase, we offer users tunability to traverse the trade-off between performance and computational cost without the need for re-training. 
While our method enables users to manage the trade-off, we also provide an additional functionality that automatically assigns pixels based on the K-means clustering algorithm which can simplify the user experience. Lastly, we introduce a post-processing technique that effectively eliminates artifacts which can arise from the distribution of computation on a pixel-wise basis.
Experiments show that our method outperforms existing patch-distributing approaches \cite{kong2021classsr, chen2022arm} in terms of the PSNR-FLOP trade-off across various SISR models \cite{dong2016accelerating, ledig2017photo, zhao2020efficient} on several benchmarks, including Test2K/4K/8K \cite{kong2021classsr} and Urban100 \cite{huang2015single}. We also compare our method with the per-image processing-based method \cite{hu2022restore}, which process images in their entirety rather than decomposing them into patches.
\section{Related Works}
\subsubsection{CNN-based SISR.} 
The evolution of deep learning in SISR begins with SRCNN \cite{dong2015image}, which introduces convolutional neural networks. VDSR \cite{kim2016accurate} deepens this approach with residual learning. SRResNet \cite{ledig2017photo} further expands the architecture using residual blocks, while EDSR \cite{lim2017enhanced} streamlines it, removing batch normalization for improved performance. RCAN \cite{zhang2018image} and RDN \cite{zhang2018residual} advance feature extraction through channel attention and dense connections, respectively. These developments have greatly improved image quality but have also raised capacity and computational costs, posing challenges for real-world applications.

\subsubsection{Lightweight SISR.}
The evolution of lightweight SISR models emphasizes efficiency in enhancing image quality. FSRCNN \cite{dong2016accelerating} starts with directly working on LR images for speed. MemNet \cite{tai2017memnet} built upon this by introducing a memory mechanism for deeper detail restoration, while CARN \cite{ahn2018fast} balances efficiency and accuracy using cascading designs. PAN \cite{zhao2020efficient} adds pixel attention for detail enhancement without heavy computational costs. LBNet \cite{gao2022lightweight} merges CNNs with transformers for high-quality SR on resource-constrained devices, and BSRN \cite{li2022blueprint} progress with a scalable approach using separable convolutions.

\subsubsection{Region-aware SISR.}
Region-aware SISR leverages the insight that high-freque-ncy regions in an image are more challenging to restore than low-frequency ones. This approach aims to enhance efficiency by reducing redundant computation in low-frequency regions. AdaDSR \cite{liu2020deep} tailors its processing depth to the image's complexity, optimizing efficiency. FAD \cite{xie2021learning} adjusts its focus based on the input's frequency characteristics, enhancing detail in critical regions while conserving effort on smoother parts. MGA \cite{hu2022restore} initially applies a global restoration to the entire image and then refines specific regions locally, guided on a predicted mask. 

Alongside, various studies have emerged focusing on efficiency in large-scale image SR. These studies decompose images into several patches and aim to enhance efficiency by dynamically allocating computational resources according to the restoration difficulty of each patch. ClassSR \cite{kong2021classsr} is the first work of this area of research: it utilizes a classifier to categorize patches into simple, medium, or hard type, and assigns them to subnets with different capacities to reduce FLOPs. However, since ClassSR employs independent subnets, it leads to a significant increase in parameter count. ARM \cite{chen2022arm} resolves the limitation by decomposing the original network into subnets that share parameters, thus no additional parameters are introduced. On the other hand, APE \cite{wang2022adaptive} uses a regressor that predicts the incremental capacity at each layer for each patch, reducing FLOPs by early patch exiting while forwarding through network layers. In this line of study, moving away from the existing patch-distributing methods, we aim to distribute computational resources on a pixel-wise, seeking additional efficiency improvements through finer granularity.
\section{Method}

\subsection{Preliminary}
Single Image Super-Resolution (SISR) is a task aimed at generating a high-resolution (HR) image from a single low-resolution (LR) input image. Within the framework of neural networks, the SISR model aims to discover a mapping function $F$ that converts a given LR image $I^{LR}$ into an HR image $I^{HR}$. It can be represented by the equation:
\begin{equation}
I^{HR} = F(I^{LR}; \theta),
\end{equation}
where $\theta$ is a set of model parameters. Typical models \cite{dong2016accelerating, ledig2017photo, lim2017enhanced, zhang2018image, zhang2018residual, ahn2018fast, zhao2020efficient, gao2022lightweight, li2022blueprint} can be decomposed into two main components: 1) a backbone $B$ that extracts features from $I^{LR}$, and 2) an upsampler $U$ that utilizes the features to reconstruct $I^{HR}$. Thus, the process can further be represented as follows:
\begin{align}
Z = B(I^{LR}; \theta_B), \quad
I^{HR} = U(Z; \theta_U).
\end{align}
Here, $\theta_B$ and $\theta_U$ are the parameters of the backbone and the upsampler respectively, and $Z$ is the extracted feature. In a convolutional neural network-based (\ie, CNN-based) upsampler, diverse operations are employed along with convolution layers to increase the resolution of the image being processed. These range from simple interpolation to more complex methods like deconvolution or sub-pixel convolution \cite{shi2016real}. Instead of using a CNN-based upsampler, one can employ a multilayer perceptron-based (\ie, MLP-based) upsampler to operate in a pixel-wise manner, which will be further described in the following section.

\begin{figure*}[t]
    \centering
    \includegraphics[width=\linewidth]{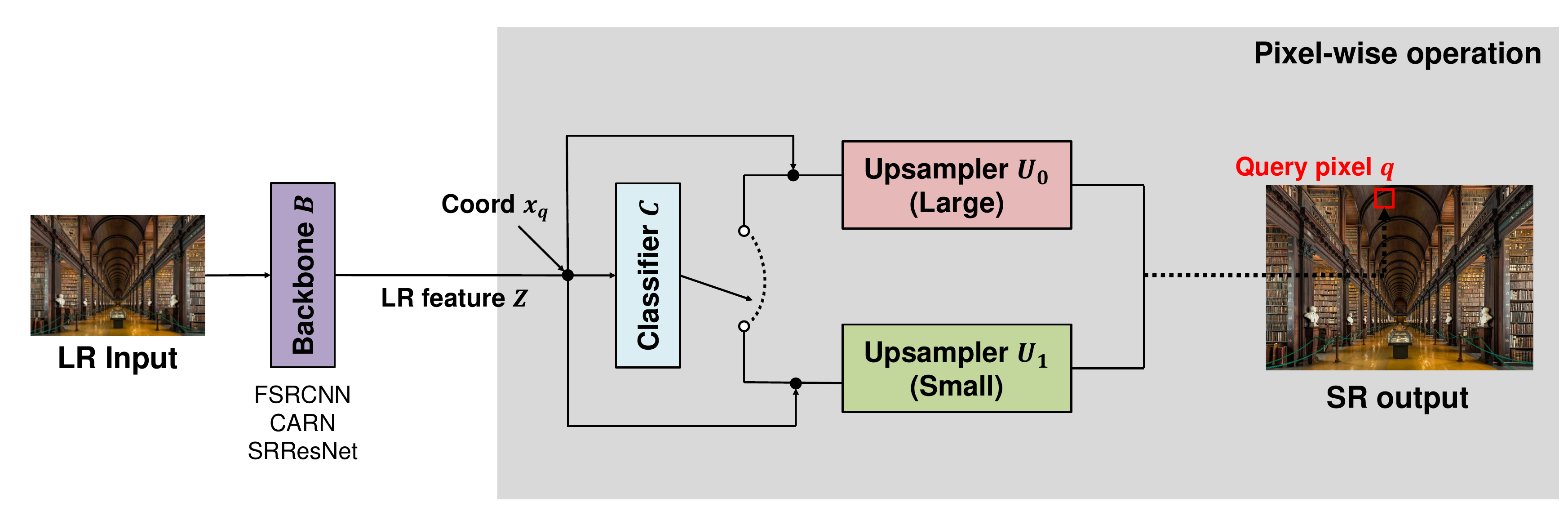}
    \caption{
    The architecture of the proposed PCSR model when the number of classes $M$ is 2.
    We denote $q$ as a single query pixel in the HR space and $x_q$ for its coordinate.
    Pixel-level probabilities obtained from the classifier are used to allocate each query pixel to a suitably-sized upsampler for the prediction of its RGB value.
    }
    \label{fig:overview}
\end{figure*}

\subsection{Network Architecture}
The overview of PCSR is shown in Fig. \ref{fig:overview}. Based on our prior discussion, a model consists of a backbone and a set of upsamplers. In addition, we employ a classifier that measures the difficulty of restoring target pixels on the HR space (\ie, query pixels). LR input image is feed-forwarded to the backbone and corresponding LR feature is generated. Then, the classifier determines the restoration difficulty for each query pixel and its output RGB value is computed through the corresponding upsampler.

\subsubsection{Backbone.}
We propose a pixel-wise computation distributing method for efficient large image SR.
It is possible to use any existing deep SR networks as our backbone to fit a desired model size.
For example, small-sized FSRCNN \cite{dong2016accelerating}, medium-sized CARN \cite{ahn2018fast}, large-sized SRResNet \cite{ledig2017photo}, and also other models can be adopted.

\subsubsection{Classifier.}
We introduce a lightweight classifier which is an MLP-based network, to obtain the probability of belonging to each upsampler (or class) in a pixel-wise manner. Given a query pixel coordinate $x_q$, our classifier assigns it to one of the corresponding upsamplers depending on the classification probability to predict its RGB value. By properly assigning easy pixels to a lighter upsampler instead of a heavier upsampler, we can save on computational resources with minimal performance drop. 

Let an LR input be $X \in \mathbb{R}^{h \times w \times 3}$, and its corresponding HR be $Y \in \mathbb{R}^{H \times W \times 3}$. And let $\{y_i\}_{i=1...HW}$ be the coordinate of each pixel within the HR $Y$ and $\{Y(y_i)\}_{i=1...HW}$ be the corresponding RGB values. Firstly, an LR feature $Z\in\mathbb{R}^{h \times w \times D}$ is calculated from the LR input using the backbone. Then, given the number of classes $M$, classification probability $p_i\in\mathbb{R}^M$ is obtained by the classifier $C$:
\begin{equation}
p_i = \sigma(C(Z, y_i; \theta_C)),
\end{equation}
where $\sigma$ is a softmax function. The MLP-based classifier operates similarly to an upsampler, with the main difference being that its output dimension is M. Please see Eq. \eqref{eq:liif1} for detailed information.

\subsubsection{Upsampler.}
We employ LIIF \cite{chen2021learning} as our upsampler, which is suitable for pixel-level processing.
We first normalize $y_i$, which is previously defined, from the HR space to map it to the coordinate $\hat{y}_i\in\mathbb{R}^2$ in the LR space. 
Given the LR feature $Z$, we denote $z_i^*\in\mathbb{R}^D$ as the nearest (by Euclidean distance) feature to the $\hat{y}_i$ and $v_i^*\in\mathbb{R}^2$ as the corresponding coordinate of that feature. Then the upsampling process is summarized as:
\begin{align}
\label{eq:liif1}
I^{SR}(y_i) = U(Z, y_i ; \theta_U) = U([z_i^*, \hat{y}_i-v_i^*]; \theta_U),
\end{align}
where $I^{SR}(y_i)\in\mathbb{R}^3$ is an RGB value at the $y_i$ and [$\cdot$] is a concatenation operation. We can obtain the final output $I^{SR}$ by querying the RGB values for every $\{y_i\}_{i=1...HW}$ and combining them (Please refer to \cite{chen2021learning} for more details of LIIF processing). In our proposed method, $M$ parallel upsamplers $\{U_0, U_1, ..., U_{M-1}\}$ can be exploited to handle a variety range of restoration difficulties (\ie from heavy to light capacity).

\subsection{Training}
During the training phase, we feed-forward a query pixel through all $M$ upsamplers and aggregate the outputs to effectively back-propagate the gradient as follows:
\begin{equation}
\label{eq:pcsr2}
\hat{Y}(y_i) = \sum_{j=0}^{M-1} p_{i,j} \times U_j(Z,y_i;\theta_{U_j}),
\end{equation}
where $\hat{Y}(y_i)\in\mathbb{R}^3$ is an RGB output at the $y_i$ and $p_{i,j}$ is the probability of that query pixel being in an upsampler $U_j$.

Then we leverage two kinds of loss functions: reconstruction loss $L_{recon}$, and average loss $L_{avg}$ which is similar one used in ClassSR \cite{kong2021classsr}. The reconstruction loss is defined as the L1 loss between the RGB values of the predicted output and the target. Here, we consider the target as the difference between the ground-truth HR patch and the bilinear upsampled LR input patch. The reason is that we want the classifier to perform the classification task well, even with a very small capacity, by emphasizing high-frequency features. Therefore, the loss can be written as:
\begin{equation}
L_{recon} = \sum_{i=1}^{HW} \lvert (Y(y_{i}) - upX(y_{i})) - \hat{Y}(y_{i}) \rvert,
\end{equation}
where $upX(y_i)$ is the RGB value of the bilinear upsampled LR input patch at the location $y_i$.
For the average loss, we encourage a uniform assignment of pixels across each class by defining the loss as:
\begin{equation}
\label{eq:loss2}
L_{avg} = \sum_{j=1}^{M} \lvert \sum_{n=1}^{N} \sum_{i=1}^{HW} p_{n,i,j} - \frac{NHW}{M} \rvert,
\end{equation}
where $p_{n,i,j}$ is probability of the $i$-th pixel of the $n$-th HR image (\ie batch dimension, with batch size $N$) being in the $j$-th class.
Here, we consider the probability for being in each class as the effective number of pixel assignments to that class. We set the target as $\frac{NHW}{M}$ because we want to allocate the same number of pixels to each class (or upsampler), out of a total of $NHW$ pixels.
Finally, total loss $L$ is defined as:
\begin{equation}
L = w_{recon} \times L_{recon} + w_{avg} \times L_{avg}.
\end{equation}

Since jointly training all modules (\ie, backbone $B$, classifier $C$, upsamplers $U_{j \in [0,M)}$) from scratch can lead to unstable training, we adopt multi-stage training strategy. Assuming that the capacity of the upsampler decreases from $U_0$ to $U_{M-1}$, the upper bound of the model's performance is determined by the backbone $B$ and the heaviest upsampler $U_0$. Thus, we initially train $\{B,U_0\}$ only using the reconstruction loss. And then, starting from $j=1$ to $j=M-1$, the following process is repeated: Firstly, freeze $\{B, U_0, ..., U_{j-1}\}$ that are trained already. Secondly, attach $U_j$ to the backbone (and also newly attach $C$ for $j=1$). Lastly, jointly train $\{U_j, C\}$ using the total loss.

\subsection{Inference}
In the inference phase of PCSR, the overall process is similar to training, but a query pixel is assigned to a unique upsampler branch based on the predicted classification probabilities. While one can allocate the pixel to the branch with the highest probability, we provide users controllability for traversing the computation-performance trade-off without re-training. To this end, FLOP count is considered in the decision-making process. We define and pre-calculate the impact of each upsampler $U_{j \in [0,M)}$ in terms of FLOPs as:
\begin{equation}
cost(U_j) = \sigma(flops(B; (h_0, w_0)) + flops(U_j; (h_0, w_0))),
\end{equation}
where $\sigma$ is the softmax function and $flops(\cdot)$ refers to FLOPs of the module, given the fixed resolution $(h_0, w_0)$\footnote{It doesn't matter whatever the values of $h_0$ and $w_0$ are, as FLOPs of the module is proportional to the input resolution. We use sufficiently small values for pre-calculating the $cost(\cdot)$ to reduce computational load.}. The branch allocation for pixel at $y_i$ is then determined as follows:
\begin{equation}
argmax_j \frac{p_{i,j}}{[cost(U_j)]^k},
\end{equation}
where $k$ is a hyperparameter and $p_{i,j}$ is the probability of that query pixel being in $U_j$, as mentioned previously. By the definition, setting lower $k$ value results in more pixels being assigned to the heavier upsamplers, minimizing performance degradation while increasing computational load. Conversely, a higher $k$ value assigns more pixels to the lighter upsamplers, accepting a reduction in performance in exchange for lower computational demand.

\subsubsection{Adaptive Decision Making (ADM).} 
While our method allows users to manage the computation-performance trade-off, we also provide an additional functionality that automatically allocates pixels based on probability values with considering statistics across the entire image. It proceeds as follows: Given $\forall p_{i,j}$ for a single input image and considering $U_{j \in [0, \lfloor(M+1)/2\rfloor)}$ as heavy upsamplers, $sum_{0 \leq j<\lfloor(M+1)/2\rfloor} p_{i,j}$ is computed to represent the restoration difficulty of that pixel, resulting in total number of $i$ values. Then we group the values into $M$ clusters using a clustering algorithm. Finally, by assigning each group to the upsamplers ranging from the heaviest $U_0$ to the lightest $U_{M-1}$ based on the its centroid value, all pixels are allocated to the appropriate upsampler. We especially employ the K-means clustering to minimize computational load. As we uniformly initialize the centroid values, the process is deterministic. We demonstrate the efficacy of ADM in the appendix.

\subsubsection{Pixel-wise Refinement.} 
Since the RGB value for each pixel is predicted by the independent upsampler, artifacts can arise when adjacent pixels are assigned to upsamplers with different capacities. To address this issue, we propose a simple solution: we again treat the lower half of the upsamplers by capacity as light upsamplers and the upper half as heavy upsamplers, performing refinement when adjacent pixels are allocated to different types of upsamplers.
To be specific, for pixels assigned to $U_{j}$ where $\lfloor(M+1)/2\rfloor \leq j < M$ (\ie, light upsamplers), if at least one neighboring pixel has been assigned to $U_{j}$ with $0 \leq j<\lfloor(M+1)/2\rfloor$ (\ie, heavy upsamplers), we replace its RGB value with the average value of the neighboring pixels (including itself) in the SR output. Our pixel-wise refinement algorithm works without needing any extra forward processing, effectively reducing artifacts with only a small amount of extra computation and having minimal effect on the overall performance.
\section{Experiments}
\label{sec:exp}

\subsection{Settings}
\subsubsection{Training.}
To ensure a fair comparison, we aligned the overall training settings to match those of ClassSR and ARM. We densely cropped DIV2K \cite{agustsson2017ntire} (from index 0001-0800) into 1.59 million 32x32 LR sub-images for training dataset and random rotation and flipping are applied for data augmentation. We adopt existing FSRCNN \cite{dong2016accelerating}, CARN \cite{ahn2018fast}, and SRResNet \cite{ledig2017photo} as backbones with their original parameters of 25K, 295K, and 1.5M respectively. Throughout all training phases for both the original models and PCSR, the batch size is 16 and the initial learning rate is set at 0.001 for FSRCNN and 0.0002 for CARN and SRResNet with cosine annealing scheduling. Adam optimizer is used. Both the original models and the initial PCSR (which includes only the backbone and the heaviest upsampler) are trained with 2,000K iterations, while subsequent stages of PCSR’s training use 500K iterations. In the initial PCSR, we fine-tuned the hidden dimension of the backbone and adjusted the MLP size of the heaviest upsampler to maintain performance parity with the original models in terms of PSNR and FLOPs. In our implementation, we simply set $M=2$ as it shows the decent performance with its simplicity, which will be verified in the Sec. \ref{ablation-studies}.

\subsubsection{Evaluation.}
We mainly evaluate our method on the Test2K/Test4K/Test8K \cite{kong2021classsr} which are downsampled from DIV8K \cite{gu2019div8k}, and the Urban100 \cite{huang2015single} which consists of much larger images than the commonly used benchmarks such as Set5 \cite{bevilacqua2012low} and Set14 \cite{yang2010image}.
For the evaluation metrics, we use PSNR (Peak Signal-to-Noise Ratio) to assess the quality of the SR images, and FLOPs (Floating Point Operations) to measure the computational efficiency. PSNR is calculated on the RGB space and FLOPs are measured on the full image.
Unless specified, the original model and our PCSR is evaluated at full resolution, while ClassSR and ARM are evaluated on an overlapped patch basis. Other evaluation protocols follow those of ClassSR and ARM. When comparing PCSR with comparison groups, pixel-wise refinement is always employed and hyperparameter $k$ is adjusted to match their performance or ADM is used.

\subsection{Main Results}
\begin{table}[t]
\centering
\caption{
The comparison of the previous patch-level methods and our pixel-level method PCSR on the large image SR benchmarks: Test2K, Test4K, Test8K, and Urban 100 with $\times$4 SR.
The lowest FLOPs values are highlighted in bold.
}
\label{tab:main}
\resizebox{\linewidth}{!}{%
\begin{tabular}{|c|c|cc|cc|}
\hline
Models            & Params. & Test2K(dB)   & GFLOPs         & Test4K(dB)  & GFLOPs       \\ \hline
FSRCNN      & 25K        & 25.69  & 45.3 (100\%)  & 26.99  & 185.3 (100\%)  \\
FSRCNN-ClassSR   & 113K       & 25.61  & 38.4 (85\%)   & 26.91  & 146.4 (79\%)   \\
FSRCNN-ARM       & 25K        & 25.61  & 35.6 (79\%)   & 26.91  & 152.9 (83\%)    \\
FSRCNN-PCSR   & 25K        & 25.61  & \textbf{8.5 (19\%})  & 26.91  &\textbf{ 32.6 (18\%)} \\ \hline
CARN        & 295K       & 26.03  & 112.0 (100\%) & 27.45  & 457.8 (100\%)   \\
CARN-ClassSR     & 645K       & 26.01  & 101.7 (91\%)  & 27.42  & 384.1 (84\%)    \\
CARN-ARM         & 295K       & 26.01  & 99.8 (89\%)  & 27.42  & 379.2 (83\%)    \\
CARN-PCSR   & 169K       & 26.01  & \textbf{64.0 (57\%)}  & 27.42   & \textbf{260.0 (58\%)}  \\ \hline
SRResNet    & 1.5M       & 26.24  & 502.9 (100\%) & 27.71  & 2056.2 (100\%) \\
SRResNet-ClassSR & 3.1M       & 26.20  & 446.7 (89\%)  & 27.66  & 1686.2 (82\%)   \\
SRResNet-ARM     & 1.5M       & 26.20  & 429.1 (85\%)  & 27.66  & 1742.2 (85\%)   \\
SRResNet-PCSR  & 1.1M       & 26.20  &\textbf{ 245.6 (49\%)}  & 27.66  & \textbf{981.0 (48\%)}  \\ \hline \hline

Models            & Params.    & Test8K(dB)         & GFLOPs           & Urban100(dB) & GFLOPs         \\ \hline
FSRCNN      & 25K        & 32.82        & 1067.8 (100\%)  & 23.05  & 19.9 (100\%)  \\
FSRCNN-ClassSR   & 113K       & 32.73        & 709.2 (66\%)    & 22.89  & 20.8 (105\%)  \\
FSRCNN-ARM       & 25K        & 32.73        & 746.7 (70\%)    & 22.89  & 19.9 (100\%)  \\
FSRCNN-PCSR   & 25K        &  32.73   & \textbf{196.6 (18\%)}  & 22.89  &\textbf{ 3.4 (17\%)}  \\ \hline
CARN        & 295K       & 33.29   & 2638.6 (100\%)  & 24.03  & 49.3 (100\%)  \\
CARN-ClassSR     & 645K       & 33.25        & 1829.9 (69\%)   & 24.00  & 51.7 (105\%)   \\
CARN-ARM         & 295K       & 33.26        & 1783.2 (68\%)   & 23.99  & 50.8 (103\%)  \\
CARN-PCSR   & 169K       & 33.25   & \textbf{1355.1 (51\%)}   & 24.00  & \textbf{29.6 (60\%)}   \\ \hline
SRResNet   & 1.5M       & 33.55   & 11850.7 (100\%) & 24.65  & 221.3 (100\%) \\
SRResNet-ClassSR & 3.1M       & 33.50        & 7996.0 (67\%)   & 24.54  & 226.5 (102\%) \\
SRResNet-ARM     & 1.5M       & 33.50        & 7865.3 (66\%)   & 24.54  & 245.2 (111\%) \\
SRResNet-PCSR  & 1.1M       & 33.52   & \textbf{5093.7 (43\%)}   & 24.54   & \textbf{124.9 (56\%)} \\ \hline
\end{tabular}
}
\end{table}

As demonstrated in Tab. \ref{tab:main}, our proposed method, PCSR, exhibits better computational efficiency compared to previous patch-based efficient SR models \cite{kong2021classsr,chen2022arm} on four benchmarks, Test2K/Test4K/Test8K, and Urban100. We assess the computational costs (FLOPs) of the existing SR models \cite{chen2022arm,kong2021classsr,hu2022restore} while ensuring their PSNR performance remain comparable.

\begin{figure}[]
\centering
\tiny
\begin{tabular}
{@{}m{0.03\linewidth}@{\hskip2pt}m{0.24\linewidth}@{\hskip2pt}m{0.14\linewidth}@{\hskip2pt}m{0.14\linewidth}@{\hskip2pt}m{0.14\linewidth}@{\hskip2pt}m{0.14\linewidth}@{\hskip2pt}m{0.14\linewidth}@{}}

&
\small\centering{Classification (Ours)} &
\small\centering{ClassSR} &
\small\centering{ARM} &
\small\centering{Ours} &
\small\centering{Backbone} &
\small\centering{GT} \tabularnewline
\midrule

\multirow{3}{*}[-6em]{\begin{subfigure}[m]{\linewidth}\caption{}\label{fig:visual_comparisons_a}\end{subfigure}} &
\multirow{3}{*}[-4em]{\includegraphics[width=\linewidth]{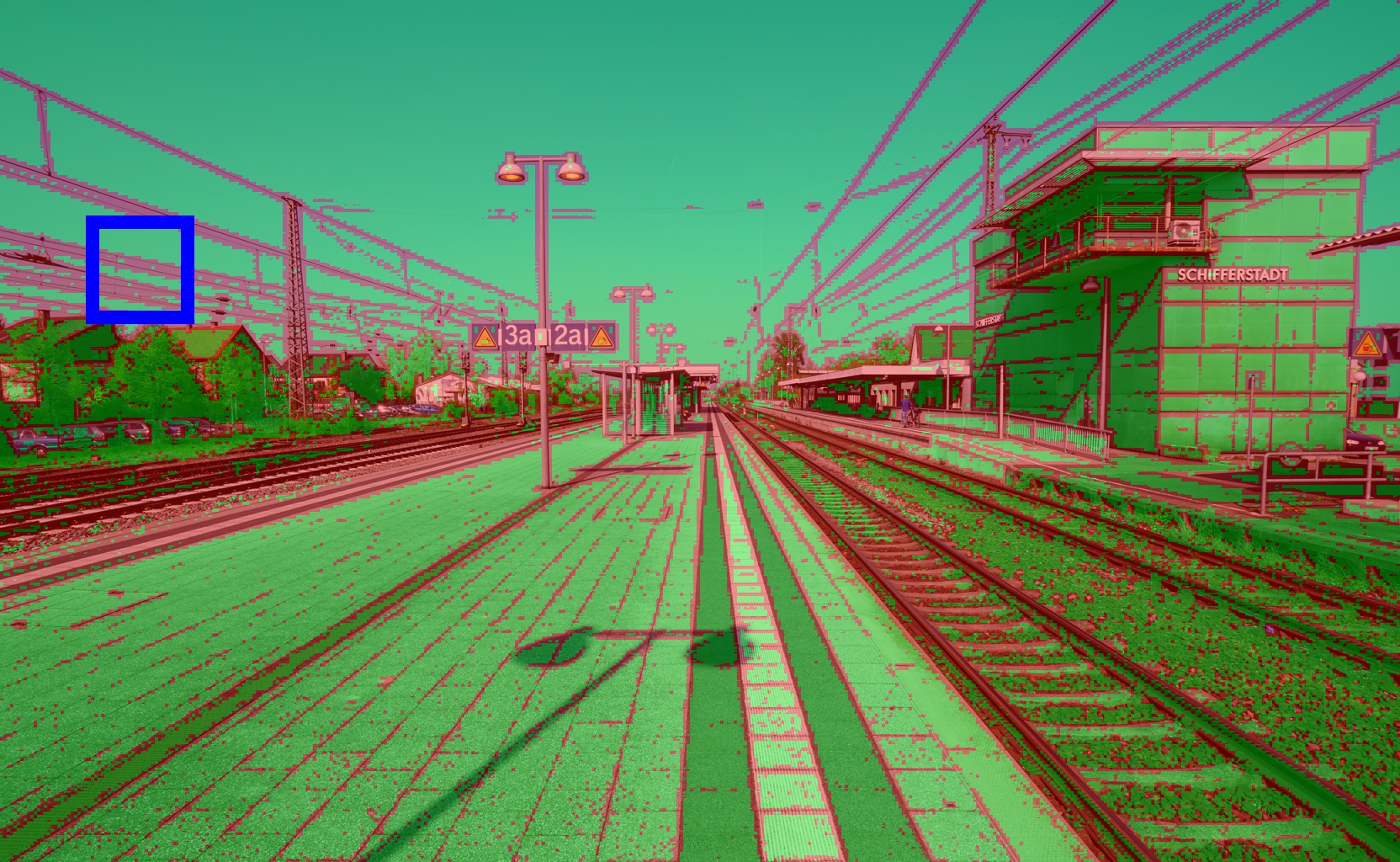}} &
\centering{26.64dB 78.2G(65\%)} &
\centering{26.66dB 77.0G(64\%)} &
\centering{26.85dB 72.4G(60\%)} &
\centering{26.87dB 120.3G(100\%)} &
\centering{} \tabularnewline
&
&
\includegraphics[width=\linewidth]{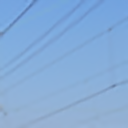} &
\includegraphics[width=\linewidth]{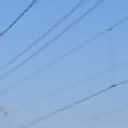} &
\includegraphics[width=\linewidth]{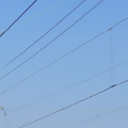} &
\includegraphics[width=\linewidth]{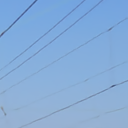} &
\includegraphics[width=\linewidth]{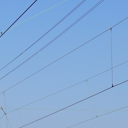} \tabularnewline
&
&
\includegraphics[width=\linewidth]{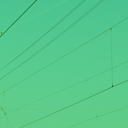} &
\includegraphics[width=\linewidth]{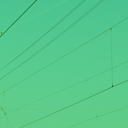} &
\includegraphics[width=\linewidth]{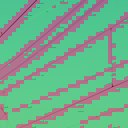} &
&
\tabularnewline
\midrule

\multirow{3}{*}[-6em]{\begin{subfigure}[m]{\linewidth}\caption{}\label{fig:visual_comparisons_b}\end{subfigure}} &
\multirow{3}{*}[-4em]{\includegraphics[width=\linewidth]{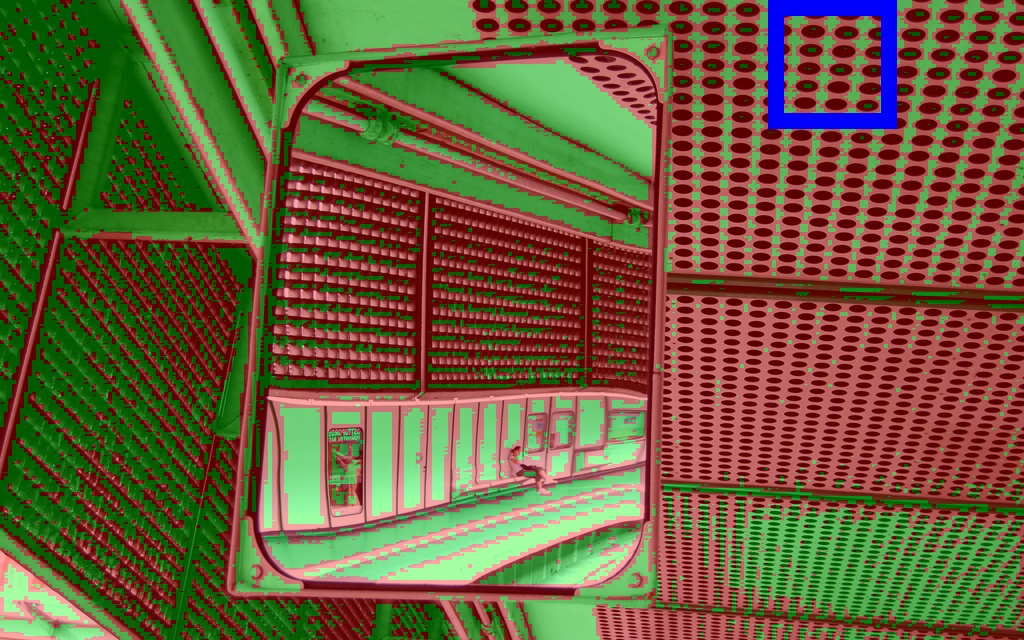}} &
\centering{21.21dB 45.0G(97\%)} &
\centering{21.10dB 45.8G(99\%)} &
\centering{21.47dB 34.5G(75\%)} &
\centering{21.31dB 46.3G(100\%)} &
\centering{} \tabularnewline
&
&
\includegraphics[width=\linewidth]{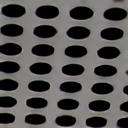} &
\includegraphics[width=\linewidth]{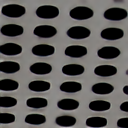} &
\includegraphics[width=\linewidth]{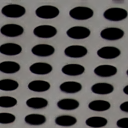} &
\includegraphics[width=\linewidth]{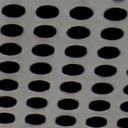} &
\includegraphics[width=\linewidth]{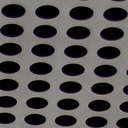} \tabularnewline
&
&
\includegraphics[width=\linewidth]{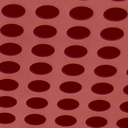} &
\includegraphics[width=\linewidth]{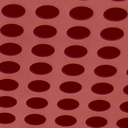} &
\includegraphics[width=\linewidth]{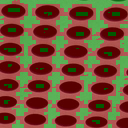} &
&
\tabularnewline
\midrule

\multirow{3}{*}[-6.5em]{\begin{subfigure}[m]{\linewidth}\caption{}\label{fig:visual_comparisons_c}\end{subfigure}} &
\multirow{3}{*}[-3.5em]{\includegraphics[width=\linewidth]{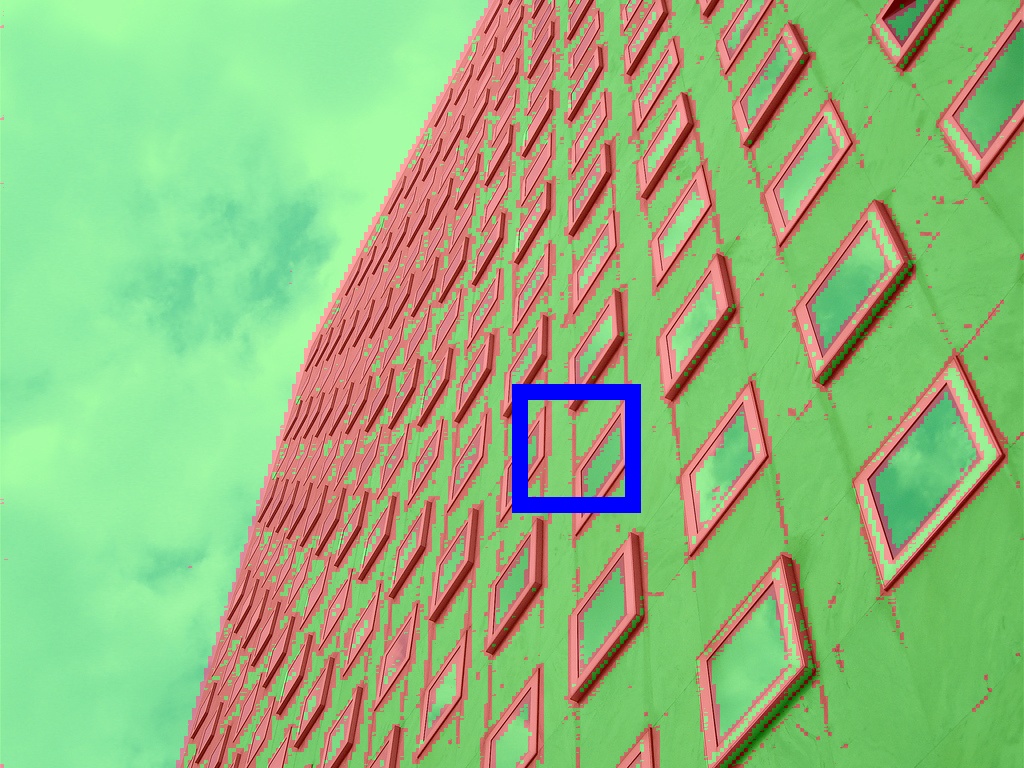}} &
\centering{25.07dB 37.3G(67\%)} &
\centering{25.07dB 35.4G(64\%)} &
\centering{25.40dB 33.3G(60\%)} &
\centering{25.37dB 55.5G(100\%)} &
\centering{} \tabularnewline
&
&
\includegraphics[width=\linewidth]{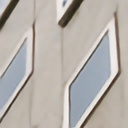} &
\includegraphics[width=\linewidth]{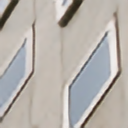} &
\includegraphics[width=\linewidth]{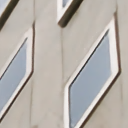} &
\includegraphics[width=\linewidth]{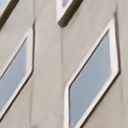} &
\includegraphics[width=\linewidth]{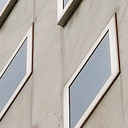} \tabularnewline
&
&
\includegraphics[width=\linewidth]{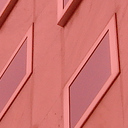} &
\includegraphics[width=\linewidth]{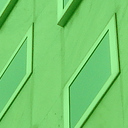} &
\includegraphics[width=\linewidth]{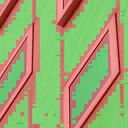} &
&
\tabularnewline
\midrule

\multirow{3}{*}[-6.5em]{\begin{subfigure}[m]{\linewidth}\caption{}\label{fig:visual_comparisons_d}\end{subfigure}} &
\multirow{3}{*}[-3.5em]{\includegraphics[width=\linewidth]{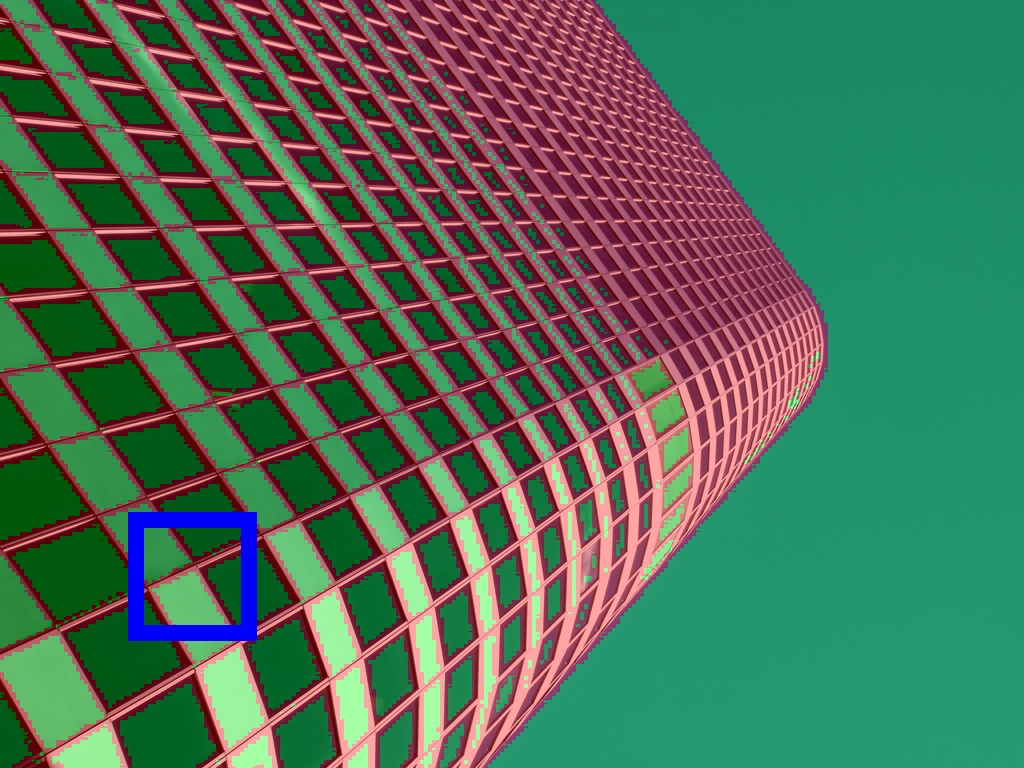}} &
\centering{25.23dB 44.3G(80\%)} &
\centering{25.18dB 37.7G(68\%)} &
\centering{25.66dB 36.6G(66\%)} &
\centering{25.43dB 55.5G(100\%)} &
\centering{} \tabularnewline
&
&
\includegraphics[width=\linewidth]{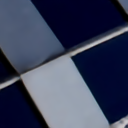} &
\includegraphics[width=\linewidth]{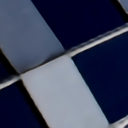} &
\includegraphics[width=\linewidth]{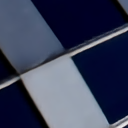} &
\includegraphics[width=\linewidth]{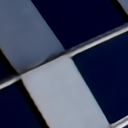} &
\includegraphics[width=\linewidth]{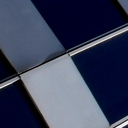} \tabularnewline
&
&
\includegraphics[width=\linewidth]{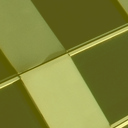} &
\includegraphics[width=\linewidth]{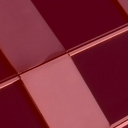} &
\includegraphics[width=\linewidth]{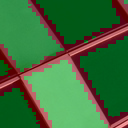} &
&
\tabularnewline

\end{tabular}
\caption{
Qualitative results of previous methods \cite{chen2022arm,kong2021classsr} and our method with $\times$4 SR. 
}
\label{fig:visual_comparisons}
\end{figure}

We also provide qualitative results with the PSNR and FLOPs of each generated image for better comparisons in Fig. \ref{fig:visual_comparisons}.
Patch-level approaches such as ClassSR and ARM fail in fine-grained restoration difficulty classification. In contrast, our method can process input image more precisely due to pixel-level classification, resulting in efficient and effective SR outputs. 
For more detailed analysis, in Fig. \ref{fig:visual_comparisons_a}, ClassSR and ARM classify the shown patch area as easy one due to the dominance of the flat region, so they fail to restore thin lines well. 
On the other hand, our method properly classifies those lines in pixel-level difficulty classification, so it recovers them well.
In Fig. \ref{fig:visual_comparisons_b}, due to over-computation by the patch-based methods, our approach demonstrates much better computational savings. This is attributed to our method's efficient distribution of computational resources, allowing us to achieve comparable or better performance while minimizing computational overhead.
In Fig. \ref{fig:visual_comparisons_c}, ClassSR waste computational resources, while ARM reduced computations excessively, resulting in inferior output quality. In contrast, our pixel-level approach enables more effective utilization of resources, leading to improved performance.

\begin{table}[t]
\centering
\caption{
The comparison of the MGA and our PCSR on Test2K, Test4K, and Urban100 with $\times$4 SR. The lowest FLOPs values are highlighted in bold.
}
\label{tab:mga}
\resizebox{\linewidth}{!}{%
\begin{tabular}{|c|c|cc|cc|cc|}
\hline
Models             & Params. & Test2K(dB) & GFLOPs      & Test4K(dB) & GFLOPs      & Urban100(dB) & GFLOPs       \\ \hline
FSRCNN             & 25K       & 25.68        & 45.3 (100\%)  & 26.98      & 185.3 (100\%) & 23.02      & 19.9 (100\%)  \\
FSRCNN-MGA         & 43K       & 25.66        & 29.2 (64\%)   & 26.94      & 101.7 (55\%)   & 23.01      & 14.6 (73\%)   \\
FSRCNN-PCSR & 25K       & 25.66        & \textbf{12.8 (28\%)} & 26.94 & \textbf{37.8 (20\%)} & 23.01 & \textbf{4.3 (22\%)} \\ \hline

SRResNet        & 1.5M       & 26.30        & 502.9 (100\%) & 27.79      & 2056.2 (100\%) & 24.87      & 221.3 (100\%) \\
SRResNet-MGA    & 2.0M       & 26.20        & 249.2 (50\%)  & 27.66      & 871.9 (42\%)  & 24.55      & 124.0 (56\%)  \\
SRResNet-PCSR & 0.9M & 26.20  & \textbf{191.0 (38\%)} & 27.66 & \textbf{755.3 (37\%)} & 24.55 & \textbf{97.3 (44\%)} \\ \hline
\end{tabular}
}
\end{table}

In Tab. \ref{tab:mga}, we further evaluate our method with the per-image processing efficient SR method, MGA \cite{hu2022restore}. To make a fair comparison, we use the same training dataset and input patch size as used in MGA and retrain our model. Even when compared to the per-image processing method, our model shows better efficiency with much fewer parameters, demonstrating its broad applicability and overall effectiveness.

\subsection{Ablation Studies}
\label{ablation-studies}

\subsubsection{Input Patch Size.}
\begin{table}[t]
\centering
\caption{
Comparison of our PCSR and ClassSR according to the patch size, on Test2K ($\times$4). To ensure a fair comparison, the original model (CARN) and our model (CARN-PCSR) are also evaluated on decomposed input patches. The LR input size is cropped to multiples of 128 without overlap to maintain consistency across patch sizes.
}
\label{tab:patch_size}
\resizebox{\linewidth}{!}{%
\begin{tabular}{|c|cc|cc|cc|cc|}
\hline
Patch Size       & \multicolumn{2}{c|}{\textbf{16}} & \multicolumn{2}{c|}{\textbf{32}} & \multicolumn{2}{c|}{\textbf{64}} & \multicolumn{2}{c|}{\textbf{128}} \\ \hline
                 & PSNR(dB)     & GFLOPs          & PSNR(dB)     & GFLOPs          & PSNR(dB)     & GFLOPs          & PSNR(dB)      & GFLOPs          \\ \hline
CARN    & 26.04        & 98.6 (100\%)     & 26.13        & 98.6 (100\%)     & 26.18        & 98.6 (100\%)     & 26.20         & 98.6 (100\%)     \\
CARN-ClassSR     & 26.03        & 66.7 (68\%)       & 26.12        & 69.8 (71\%)       & 26.16        & 72.5 (74\%)       & 26.17         & 75.8 (77\%)       \\
CARN-PCSR & 26.03        & \textbf{61.1 (62\%)}       & 26.12        & \textbf{60.3 (61\%)}       & 26.16        & \textbf{56.9 (58\%)}       & 26.17         & \textbf{54.5 (55\%)}       \\ \hline
\end{tabular}
}
\end{table}

As shown in Tab. \ref{tab:patch_size}, our experiments demonstrate that efficiency of the patch-distributing method \cite{kong2021classsr} decreases as the size of the patch increases.
This decline occurs because larger patches are more likely to contain a mix of easy and hard regions at the pixel level, making precise prediction of patch difficulty more challenging.
In contrast to the patch-level approach, our method employs a pixel-level approach, allowing any patch sizes without computational efficiency decline.
Our method is more efficient than the patch-level approach at all patch sizes, with the gap becoming more pronounced as the patch size increases.

\subsubsection {Impact of the number of classes.}
\begin{table}[t]
\centering
\caption{
Comparison depending on the number of classes $M$ with $\times$4 SR.
}
\label{tab:num_class}
\resizebox{\linewidth}{!}{%
\begin{tabular}{|c|c|cc|cc|cc|}
\hline
Models             & Params. & Test2K(dB) & GFLOPs      & Test4K(dB) & GFLOPs      & Urban100(dB) & GFLOPs       \\ \hline
CARN             & 295K       & 26.03        & 112.0 (100\%)  & 27.45      & 457.8 (100\%) & 24.03      & 49.3 (100\%)  \\
CARN-PCSR-2class         & 169K       & 26.01        & 64.0 (57\%)   & 27.42      & 260.0 (58\%)   & 24.00      & 29.6 (60\%)   \\
CARN-PCSR-3class & 181K       & 26.01        & 62.4 (56\%) & 27.42 & 245.1 (54\%) & 24.00 & 28.6 (58\%) \\ \hline
\end{tabular}
}
\end{table}

In Table \ref{tab:num_class}, we explore the impact of the number of classes on the efficiency of PCSR by comparing cases with $M$=2 and $M$=3. 
While both scenarios exhibit high efficiency compared to the original model, the case with fewer classes has minimal impact on efficiency while using fewer parameters. Therefore, for simplicity, we choose $M$=2.

\subsubsection{Multi-scale SR.}
\begin{table}[t]
\centering
\caption{
Comparison of multi-scale PCSR and ARM on Test2K. Our model (CARN-PCSR) is retrained in a multi-scale training setting with a scale range of [2,4].
}
\label{tab:multi_scale}
\resizebox{\linewidth}{!}{%
\begin{tabular}{|c|c|ccc|ccc|ccc|}
\hline
              &                  & \multicolumn{3}{c|}{x2}          & \multicolumn{3}{c|}{x4}          & \multicolumn{3}{c|}{x8}    \\ \hline
Models        & Total Params. & Params. & PSNR & FLOPs & Params. & PSNR & FLOPs & Params. & PSNR & FLOPs \\ \hline
CARN-original & \textbf{885K}     & 258K        & 30.79dB    & 335G      & 295K        & 26.03dB    & 112G      & 332K        & 23.51dB    & 57G       \\
CARN-ARM      & \textbf{885K}     & 258K        & 30.57dB    & 181G      & 295K        & 25.85dB    & 60G       & 332K        & 23.17dB    & 31G       \\
CARN-PCSR     & \textbf{169K}     & 169K        & 30.57dB    & 233G      & 169K        & 25.85dB    & 56G       & 169K        & 23.48dB    & 31G       \\ \hline
\end{tabular}
}
\end{table}
By leveraging LIIF \cite{chen2021learning} as our upsampler, our model inherently benefits from LIIF's key feature of multi-scale SR. It allows us to maintain efficiency that only a single model is required to accommodate diverse scale factors, unlike other methods which necessitate individual models for each scale factor. We demonstrate this advantage of LIIF-based upsampling in Tab. \ref{tab:multi_scale}. Furthermore, our model can extend to arbitrary-scale SR, including non-integer scales, a capability not achievable with conventional patch-based approaches.

\subsubsection{Pixel-wise Refinement.}
\begin{figure*}[t]
    \centering
    \includegraphics[width=\linewidth]{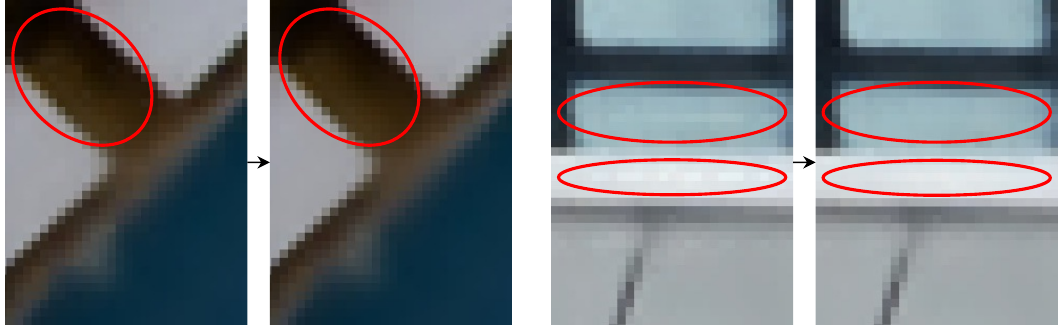}
    \caption{
    Visualization of the artifact reduction by the pixel-wise refinement.
    }
    \label{fig:artifact}
\end{figure*}
In a patch-level approach, using individual models based on patch-wise difficulties can result in artifacts when adjacent areas are assigned to different models. This issue can be mitigated by employing patch overlapping, where overlapped areas are averaged with multiple patch-level SR outputs. However, this solution harms computational efficiency by increasing the number of patches per image.
Similarly, using upsamplers based on pixel-wise difficulties can cause artifacts if neighboring pixels are assigned to different upsamplers.
Our pixel-wise refinement algorithm does not require any additional forward processing, allowing artifacts to be effectively mitigated with minor additional computations and minimal impact on performance.
Fig. \ref{fig:artifact} illustrates the efficacy of our simple yet effective pixel-wise refinement algorithm.
\section{Limitation and Future Works}
Our PCSR dynamically allocates resources based on the restoration difficulty of each pixel, thus persuing further efficiency improvements through finer granularity.
Nevertheless, a limitation exists: since our classifier operates based on LR features from backbone, the lower bound of PCSR’s FLOPs is determined by the size of the backbone. This can lead to unnecessary computation for images with predominantly flat regions. 
To mitigate this, we plan to have the classifier work on the backbone's earlier layers or use a lookup table for straightforward pixel processing through bilinear interpolation from the LR input, significantly reducing computational costs compared to neural network processing.
Additionally, for future works, applying the PCSR to generative models to enhance efficiency, as well as integrating it with techniques such as model compression, pruning, and quantization, presents promising opportunities.
\section{Conclusion}
This paper introduces the Pixel-level Classifier for Single Image Super-Resolution (PCSR), a novel approach to efficient SR for large images. 
Unlike existing patch-distributing methods, PCSR allocates computational resources at the pixel level, addressing varying restoration difficulties and reducing redundant computations with finer granularity.
It also offers tunability during inference, balancing performance and computational cost without re-training. Additionally, an automatic pixel assignment using K-means clustering and a post-processing technique to remove artifacts are also provided.
Experiments show that PCSR outperforms existing methods in the PSNR-FLOP trade-off across various SISR models and benchmarks.
We believe our proposed method facilitates the practicality and accessibility of large image SR for real-world applications.
\section*{Acknowledgement}
This research was supported and funded by Artificial Intelligence Graduate School Program under Grant (2020-0-01361), the National Research Foundation of Korea(NRF) grant funded by the Korea government (MSIT) (NRF-2022R1A2C2004509), and Samsung Electronics Co., Ltd. (Mobile eXperience Business).

%
%
\bibliographystyle{splncs04}
\bibliography{egbib}

\appendix


\title{Appendix for \\ 
Accelerating Image Super-Resolution \\
Networks with Pixel-Level Classification} 

\titlerunning{Accelerating Image SR \\
Networks with Pixel-Level Classification}

\author{Jinho Jeong\inst{1}\orcidlink{0009-0004-0947-0508} \and
Jinwoo Kim\inst{1}\orcidlink{0009-0001-3250-1788} \and
Younghyun Jo\inst{2}\orcidlink{0000-0002-8530-9802} \and
Seon Joo Kim\inst{1}\orcidlink{0000-0001-8512-216X}}

\authorrunning{J. Jeong et al.}

\institute{Yonsei University \and Samsung Advanced Institute of Technology}

\maketitle

\section{Adaptive Decision Making (ADM)}
During the inference phase of our PCSR, we provide additional functionality: Adaptive Decision Making (ADM), which automatically assigns pixels to proper-sized branches. 
While a simple approach is to allocate the pixel to the branch with the highest probability, ADM differs by taking into account statistical values of probabilities across the entire image. 
The value for each $i$-th pixel in the image initially determined through $sum_{0 \leq j<\lfloor(M+1)/2\rfloor} p_{i,j}$ to represent the restoration difficulty of that pixel, considering $U_{j \in [0, \lfloor(M+1)/2\rfloor)}$ as heavy upsamplers.
Subsequently, these difficulty values are used to perform k-means clustering with M clusters and each clusters are assigned to the corresponding branch.

\begin{figure*}[t]
    \centering
    \includegraphics[width=0.7\linewidth]{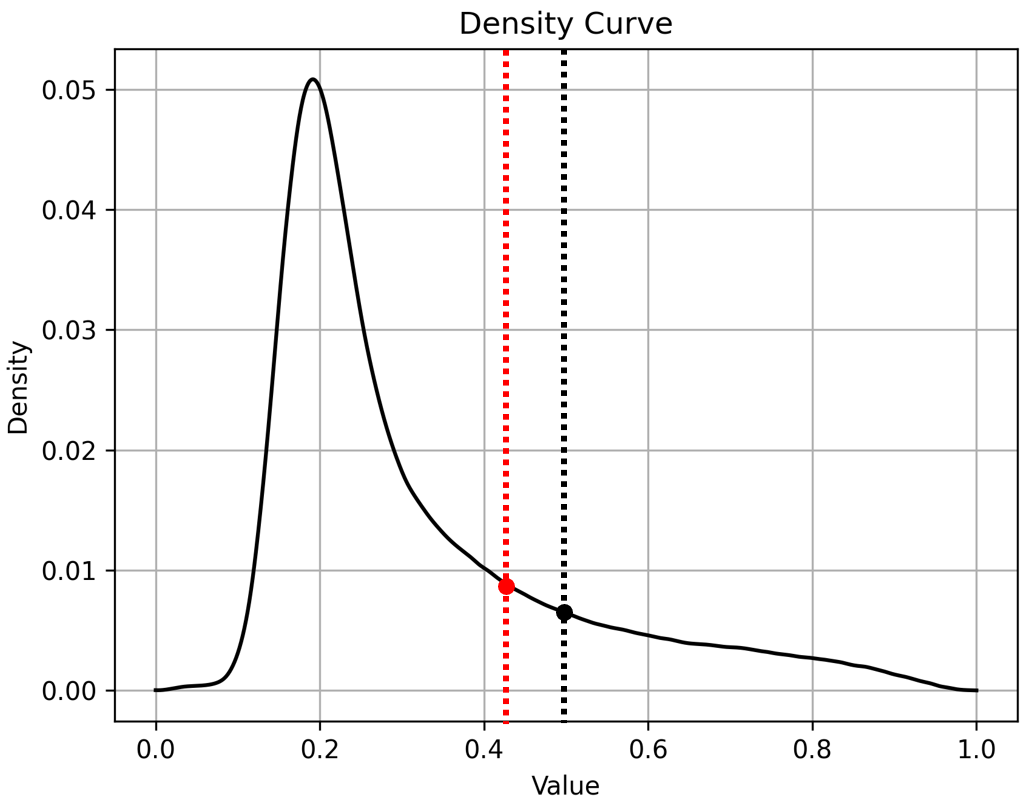}
    \caption{
    Difficulty density curve for the image “0855” (DIV2K) with $M$=2 on $\times$4.
    The range of values are divided into 100 bins, with density calculated as the count of values per bin divided by the total value count. The density, associated with each bin's center, is interpolated to form a smooth curve.
    Each dotted line indicates threshold for assigning pixels: pixels left of a line go to the light upsampler, those to the right to the heavy upsampler. 
    The black dotted line represents a threshold ($=0.5$) of simple approach (\ie, allocating pixels to the upsampler with the highest probability), while red dotted line indicates an adaptively determined threshold by ADM.
    }
    \label{fig:adm}
\end{figure*}

We show the potential of ADM through Fig. \ref{fig:adm}. 
While the simple approach fixes the threshold at 0.5 regardless of images, ADM adaptively forms the threshold at the point where the density of difficulty starts to sufficiently decrease by clustering areas with high value density.
That is, ADM avoids regions where even minor variations in the threshold could lead to sensitive changes in pixel allocation. It instead allows the threshold to be established in a section that remains stable against these variations, ensuring a more consistent allocation.
Additionally, since only a few iterations (about 2-7 iters per image) are required for clustering to converge, the additional overhead by ADM is negligible.

\section{More Experiments}

\subsection {Results on other benchmarks}
\begin{table}[t]
\centering
\caption{\footnotesize PSNR and FLOPs for additional benchmarks on $\times$4.}
\label{tab:additional_datasets}
{
\begin{tabular}{|c|cc|cc|cc|}
\hline
Model    & Set14(dB) & FLOPs  & B100(dB) & FLOPs & Manga109(dB) & FLOPs  \\ \hline
CARN     & 26.52     & 13.53G & 26.37    & 7.86G & 27.93        & 64.01G \\
+ClassSR & 26.48     & 13.25G & 26.32    & 6.88G & 27.86        & 68.04G \\
+ARM     & 26.48     & 14.12G & 26.32    & 6.56G & 27.88        & 69.79G \\
\textbf{+PCSR}    & 26.48    & \textbf{8.14G}  & 26.32    & \textbf{4.19G} & 27.86        & \textbf{40.66G} \\ \hline
\end{tabular}
}
\end{table}
We provide results for other benchmarks including Set14, B100, and Manga109. As shown in Tab. \ref{tab:additional_datasets}, our method is still efficient even for images of moderate size, compared to patch-based methods.

\subsection {Running Time Comparison}
\begin{table}[t]
\centering
\caption{\footnotesize Comparison of running time per image on $\times$4, when the performance of ARM and PCSR is set to be the same as ClassSR.}
\label{tab:runtime}
\centering
{
\begin{tabular}{|c|ccc|}
\hline
\multicolumn{1}{|c|}{Model(CARN)} & Urban100 & Test2K & Test4K  \\ \hline
+ClassSR           & 1994ms   & 4595ms & 19072ms \\
+ARM               & 518ms    & 1069ms & 4608ms  \\
\textbf{+PCSR}             & \textbf{45ms}        & \textbf{62ms}     & \textbf{203ms}       \\ \hline
\end{tabular}
}
\end{table}
Tab. \ref{tab:runtime} compares the running time between the patch-based methods and our method. Although the running time of ours is much faster, note that all methods primarily aim to reduce FLOPs, and the implementations are not fully optimized for the running time. We will look into more efficient implementation.
\section{More Ablation Studies}

\subsection {Impact of the condition for pixel-wise refinement}
\begin{table}[t]
\centering
\caption{
Variation in PCSR performance on Test2K ($\times$4) depending on the condition for pixel-wise refinement. Here, "\#h" denotes the threshold number of neighboring pixels allocated to heavy upsamplers required around a pixel to trigger its replacement. \#h=9 can be considered as the performance where no refinement is performed.
}
\label{tab:refinement}
\begin{tabular}{|c|cccccc|}
\hline
Model     & \multicolumn{6}{c|}{CARN-PCSR}                                                                                                                                        \\ \hline
\#h       & \multicolumn{1}{c|}{0}      & \multicolumn{1}{c|}{2}               & \multicolumn{1}{c|}{4}      & \multicolumn{1}{c|}{6}      & \multicolumn{1}{c|}{8}      & 9    \\ \hline
PSNR (dB) & \multicolumn{1}{c|}{25.995} & \multicolumn{1}{c|}{26.011} & \multicolumn{1}{c|}{26.016} & \multicolumn{1}{c|}{26.021} & \multicolumn{1}{c|}{26.022} & 26.022 \\ \hline
\end{tabular}
\end{table}
Pixel-wise refinement is designed to minimize artifacts by adjusting the RGB values of pixels assigned to light upsamplers to the average RGB value of their neighbors if any adjacent pixels are assigned to heavy upsamplers. 
We investigate how many neighboring pixels should be allocated to heavy upsamplers to effectively reduce artifacts while maintaining performance, as shown in Tab. \ref{tab:refinement}. 

Interestingly, we observe negligible performance degradation for any condition, even when all the pixels assigned to light upsamplers are replaced regardless of the status of neighboring pixels (\ie, \#h=0).
According to the table, while there is a slight decrease in performance when at least one neighboring pixel is allocated to heavy upsamplers (\ie, \#h=1), this condition results in a greater number of replaced pixels, which is beneficial for artifact removal. Therefore, we choose \#h=1 and always activate refinement in our evaluation.

\subsection {Impact of the LIIF Upsampler}
\begin{table}[]
\centering
\caption{\footnotesize Comparison between pixel-shuffle upsampler and LIIF upsampler on $\times$4. \textbf{MAX} denotes maximum PSNR and FLOPs by our method.}
\label{tab:different_upsampler}
{
\begin{tabular}{|c|cc|cc|cc|}
\hline
Model      & Test2K(dB) & FLOPs  & Test4K(dB) & FLOPs   & Urban100(dB) & FLOPs  \\ \hline
FSRCNN     & 25.69      & 45.3G  & 26.99      & 185.3G  & 23.05      & 19.9G  \\
+PCSR(MAX) & 25.69      & 44.5G  & 27.01      & 181.8G  & 23.27      & 19.6G  \\ \hline
CARN       & 26.03      & 112.0G & 27.45      & 457.8G  & 24.03      & 49.3G  \\
+PCSR(MAX) & 26.05      & 114.4G & 27.47      & 467.7G  & 24.09      & 50.3G  \\ \hline
SRResNet   & 26.24      & 502.9G & 27.71      & 2056.2G & 24.65      & 221.3G \\
+PCSR(MAX)    & 26.24      & 507.9G & 27.71      & 2076.6G & 24.63      & 223.5G \\ \hline
\end{tabular}
}
\end{table}
We compare between LIIF-based and CNN (or pixelshuffle)-based upsamplers in Tab. \ref{tab:different_upsampler}. The performance of the model can be higher with the LIIF upsampler than the original (\eg, FSRCNN, CARN), but for SRResNet, the performance is same or even lower than the original. Hence, we argue that the adoption of the LIIF does not guarantee the higher performance.
\section{Effectiveness on the Recent Lightweight Model}
To further demonstrate PCSR's broad applicability and efficiency, we apply our PCSR method to the recent lightweight model, BSRN \cite{li2022blueprint}.
BSRN is the model that won first place in the model complexity track of the NTIRE 2022 Efficient SR Challenge, utilizing separable convolutions to enhance its scalability. 
The result is shown in Tab. \ref{tab:bsrn}, illustrating that PCSR achieves performance comparable on several large image-based benchmarks while using fewer FLOPs. This highlights the versatility and effectiveness of our approach.
\section{More Visual Comparisons}

\begin{figure}[]
\centering
\tiny
\begin{tabular}
{@{}m{0.03\linewidth}@{\hskip2pt}m{0.24\linewidth}@{\hskip2pt}m{0.14\linewidth}@{\hskip2pt}m{0.14\linewidth}@{\hskip2pt}m{0.14\linewidth}@{\hskip2pt}m{0.14\linewidth}@{\hskip2pt}m{0.14\linewidth}@{}}

&
\small\centering{Classification (Ours)} &
\small\centering{ClassSR} &
\small\centering{ARM} &
\small\centering{Ours} &
\small\centering{Backbone} &
\small\centering{GT} \tabularnewline
\midrule

\multirow{3}{*}[-6em]{\begin{subfigure}[m]{\linewidth}\caption{}\label{fig:visual_comparisons_2k_a}\end{subfigure}} &
\multirow{3}{*}[-4em]{\includegraphics[width=\linewidth]{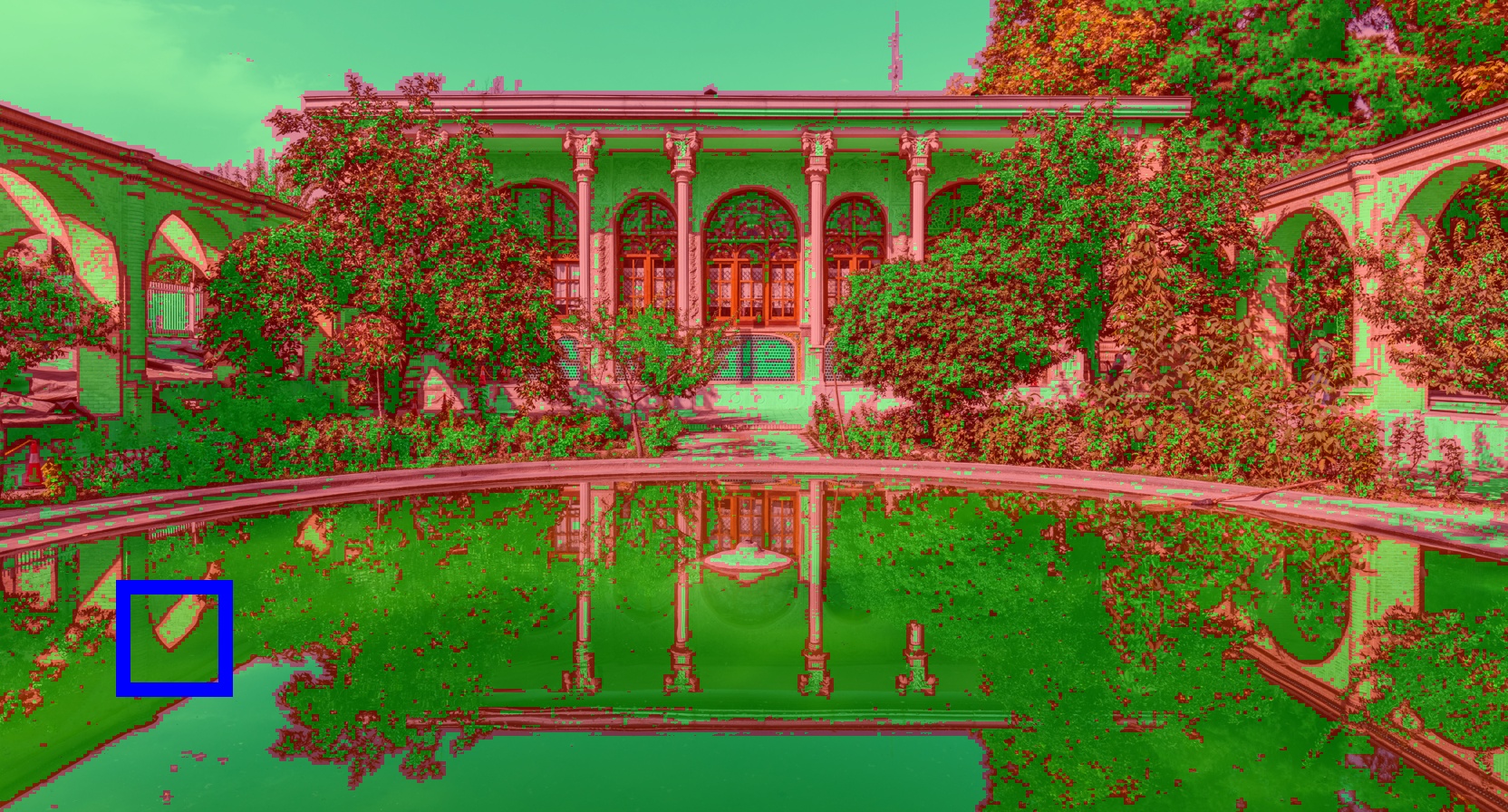}} &
\centering{21.89dB 87.4G(83\%)} &
\centering{21.89dB 78.5G(75\%)} &
\centering{21.90dB 67.5G(64\%)} &
\centering{21.92dB 105.2G(100\%)} &
\centering{} \tabularnewline
&
&
\includegraphics[width=\linewidth]{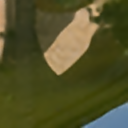} &
\includegraphics[width=\linewidth]{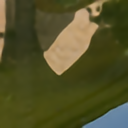} &
\includegraphics[width=\linewidth]{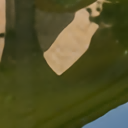} &
\includegraphics[width=\linewidth]{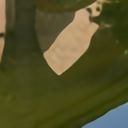} &
\includegraphics[width=\linewidth]{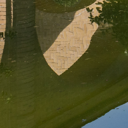} \tabularnewline
&
&
\includegraphics[width=\linewidth]{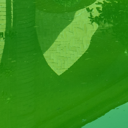} &
\includegraphics[width=\linewidth]{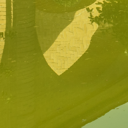} &
\includegraphics[width=\linewidth]{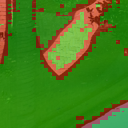} &
&
\tabularnewline
\midrule

\multirow{3}{*}[-6em]{\begin{subfigure}[m]{\linewidth}\caption{}\label{fig:visual_comparisons_2k_b}\end{subfigure}} &
\multirow{3}{*}[-4em]{\includegraphics[width=\linewidth]{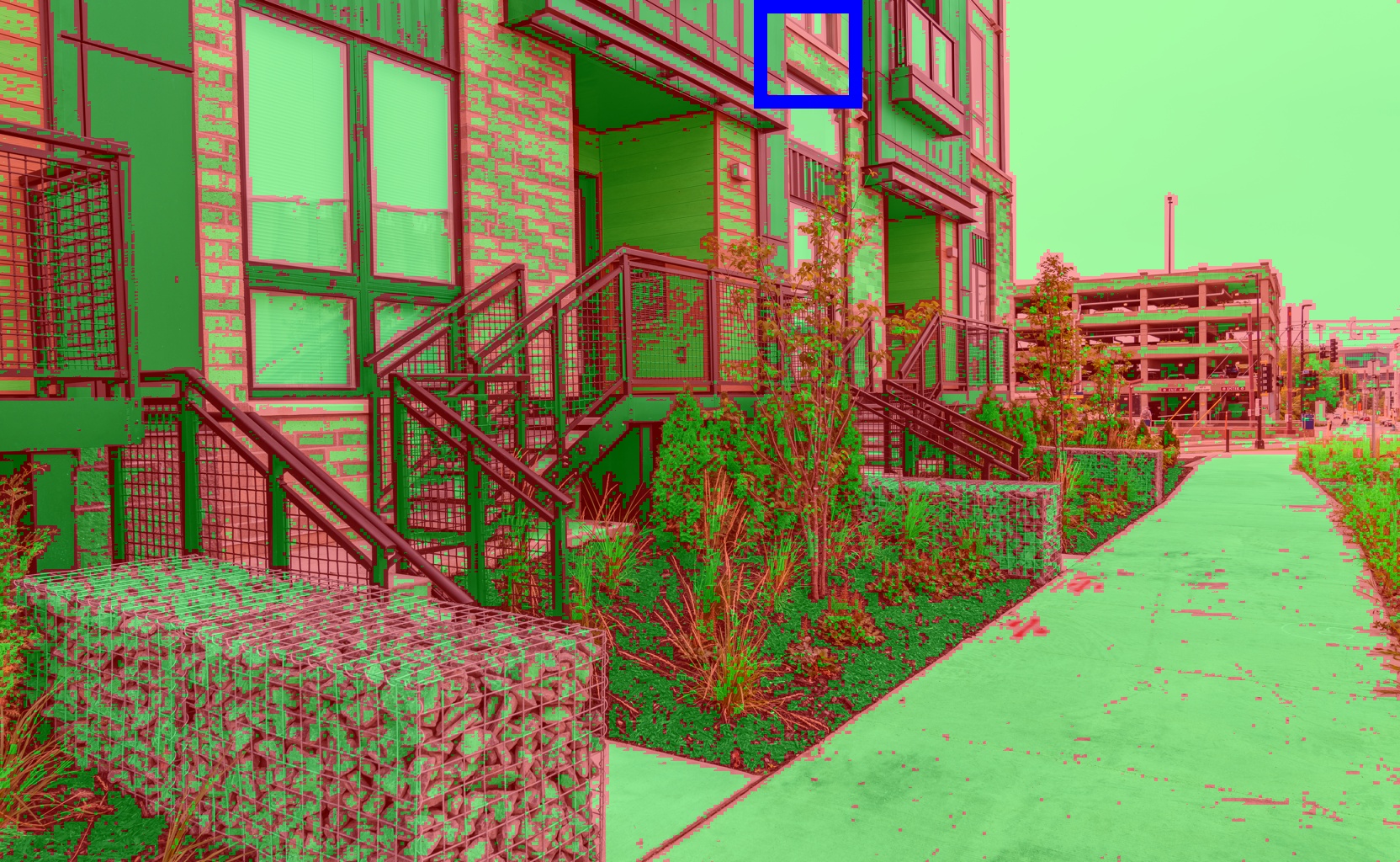}} &
\centering{23.38dB 101.4G(84\%)} &
\centering{23.26dB 82.3G(68\%)} &
\centering{23.42dB 75.3G(63\%)} &
\centering{23.45dB 120.3G(100\%)} &
\centering{} \tabularnewline
&
&
\includegraphics[width=\linewidth]{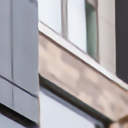} &
\includegraphics[width=\linewidth]{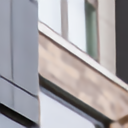} &
\includegraphics[width=\linewidth]{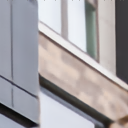} &
\includegraphics[width=\linewidth]{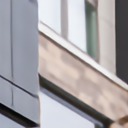} &
\includegraphics[width=\linewidth]{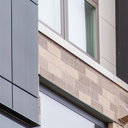} \tabularnewline
&
&
\includegraphics[width=\linewidth]{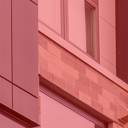} &
\includegraphics[width=\linewidth]{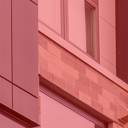} &
\includegraphics[width=\linewidth]{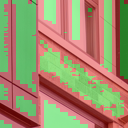} &
&
\tabularnewline
\midrule

\multirow{3}{*}[-6em]{\begin{subfigure}[m]{\linewidth}\caption{}\label{fig:visual_comparisons_2k_c}\end{subfigure}} &
\multirow{3}{*}[-4em]{\includegraphics[width=\linewidth]{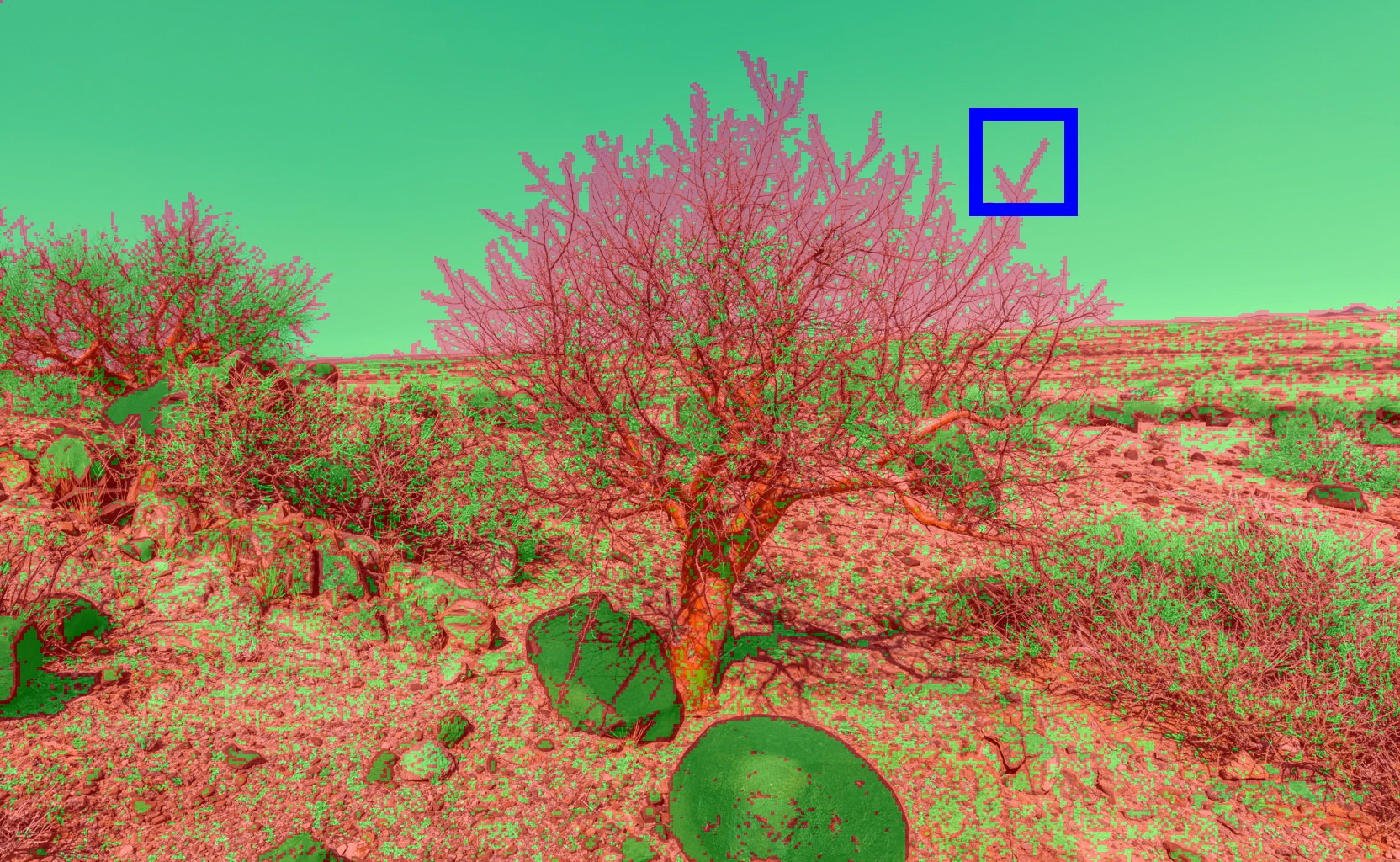}} &
\centering{21.68dB 94.8G(79\%)} &
\centering{21.68dB 86.6G(72\%)} &
\centering{21.69dB 74.4G(62\%)} &
\centering{21.71dB 120.3G(100\%)} &
\centering{} \tabularnewline
&
&
\includegraphics[width=\linewidth]{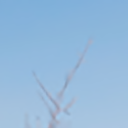} &
\includegraphics[width=\linewidth]{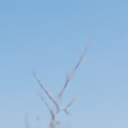} &
\includegraphics[width=\linewidth]{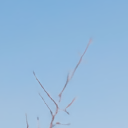} &
\includegraphics[width=\linewidth]{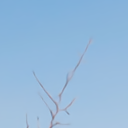} &
\includegraphics[width=\linewidth]{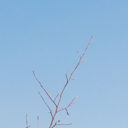} \tabularnewline
&
&
\includegraphics[width=\linewidth]{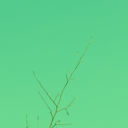} &
\includegraphics[width=\linewidth]{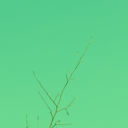} &
\includegraphics[width=\linewidth]{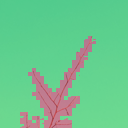} &
&
\tabularnewline
\midrule

\multirow{3}{*}[-6em]{\begin{subfigure}[m]{\linewidth}\caption{}\label{fig:visual_comparisons_2k_d}\end{subfigure}} &
\multirow{3}{*}[0.2em]{\hspace{2.2em}\includegraphics[width=0.6\linewidth]{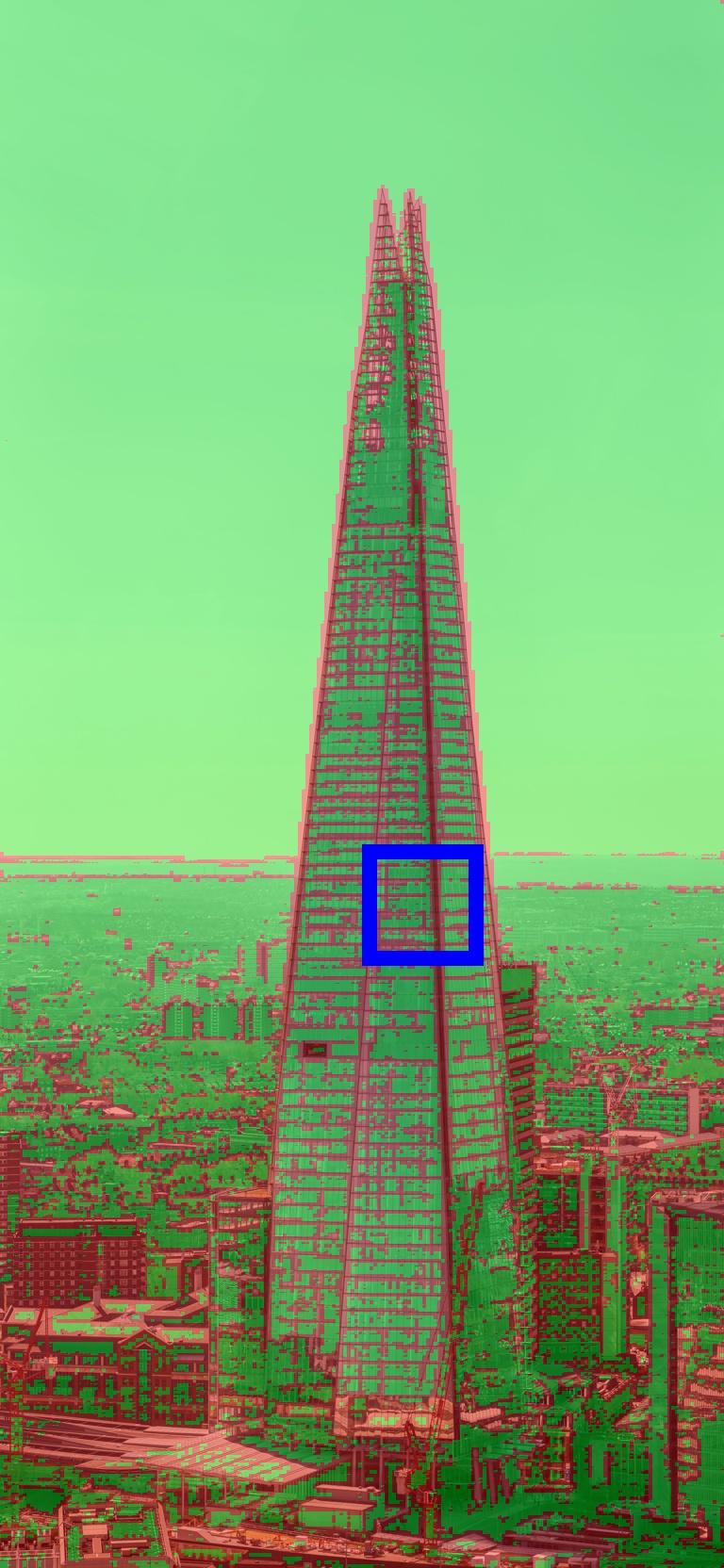}} &
\centering{28.95dB 53.3G(59\%)} &
\centering{28.99dB 42.0G(47\%)} &
\centering{29.11dB 52.8G(58\%)} &
\centering{29.12dB 90.2G(100\%)} &
\centering{} \tabularnewline
&
&
\includegraphics[width=\linewidth]{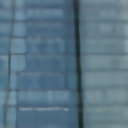} &
\includegraphics[width=\linewidth]{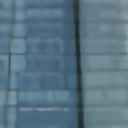} &
\includegraphics[width=\linewidth]{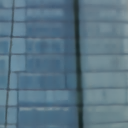} &
\includegraphics[width=\linewidth]{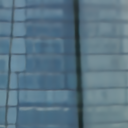} &
\includegraphics[width=\linewidth]{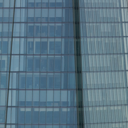} \tabularnewline
&
&
\includegraphics[width=\linewidth]{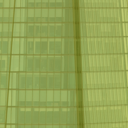} &
\includegraphics[width=\linewidth]{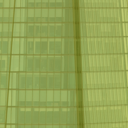} &
\includegraphics[width=\linewidth]{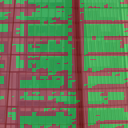} &
&
\tabularnewline

\end{tabular}
\caption{
Qualitative results of the previous methods \cite{chen2022arm,kong2021classsr} and our method with $\times$4 SR on Test2K.
}
\label{fig:visual_comparisons_test2k}
\end{figure}

\begin{figure}[]
\centering
\tiny
\begin{tabular}
{@{}m{0.03\linewidth}@{\hskip2pt}m{0.24\linewidth}@{\hskip2pt}m{0.14\linewidth}@{\hskip2pt}m{0.14\linewidth}@{\hskip2pt}m{0.14\linewidth}@{\hskip2pt}m{0.14\linewidth}@{\hskip2pt}m{0.14\linewidth}@{}}

&
\small\centering{Classification (Ours)} &
\small\centering{ClassSR} &
\small\centering{ARM} &
\small\centering{Ours} &
\small\centering{Backbone} &
\small\centering{GT} \tabularnewline
\midrule

\multirow{3}{*}[-6em]{\begin{subfigure}[m]{\linewidth}\caption{}\label{fig:visual_comparisons_4k_a}\end{subfigure}} &
\multirow{3}{*}[-4em]{\includegraphics[width=\linewidth]{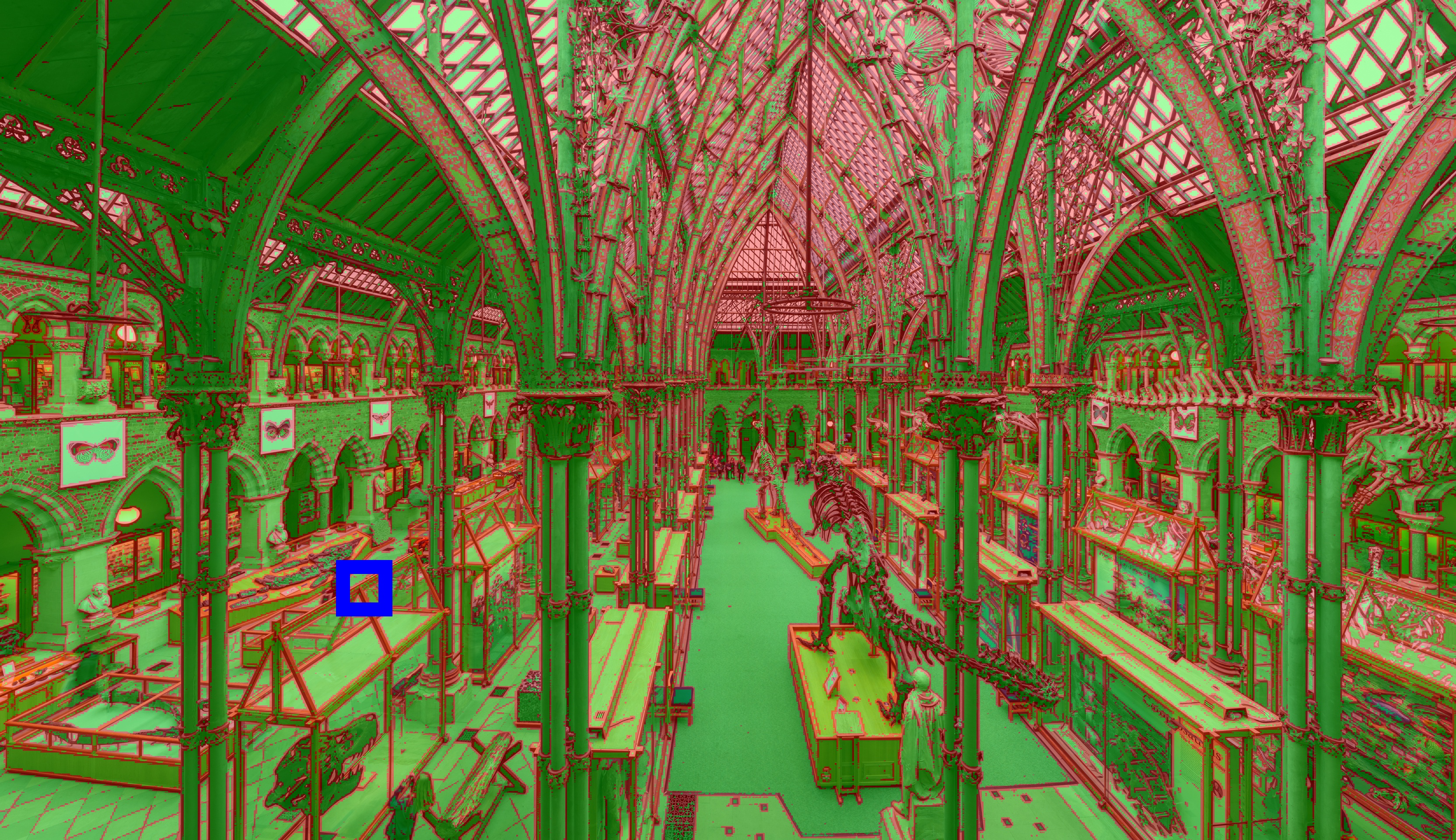}} &
\centering{25.96dB 379.0G(84\%)} &
\centering{25.94dB 393.3G(87\%)} &
\centering{26.10dB 298.5G(66\%)} &
\centering{26.13dB 451.0G(100\%)} &
\centering{} \tabularnewline
&
&
\includegraphics[width=\linewidth]{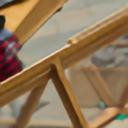} &
\includegraphics[width=\linewidth]{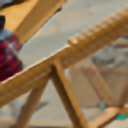} &
\includegraphics[width=\linewidth]{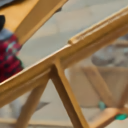} &
\includegraphics[width=\linewidth]{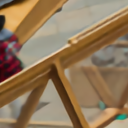} &
\includegraphics[width=\linewidth]{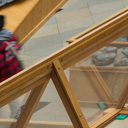} \tabularnewline
&
&
\includegraphics[width=\linewidth]{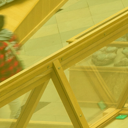} &
\includegraphics[width=\linewidth]{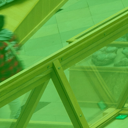} &
\includegraphics[width=\linewidth]{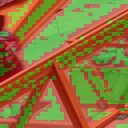} &
&
\tabularnewline
\midrule

\multirow{3}{*}[-6em]{\begin{subfigure}[m]{\linewidth}\caption{}\label{fig:visual_comparisons_4k_b}\end{subfigure}} &
\multirow{3}{*}[-4em]{\includegraphics[width=\linewidth]{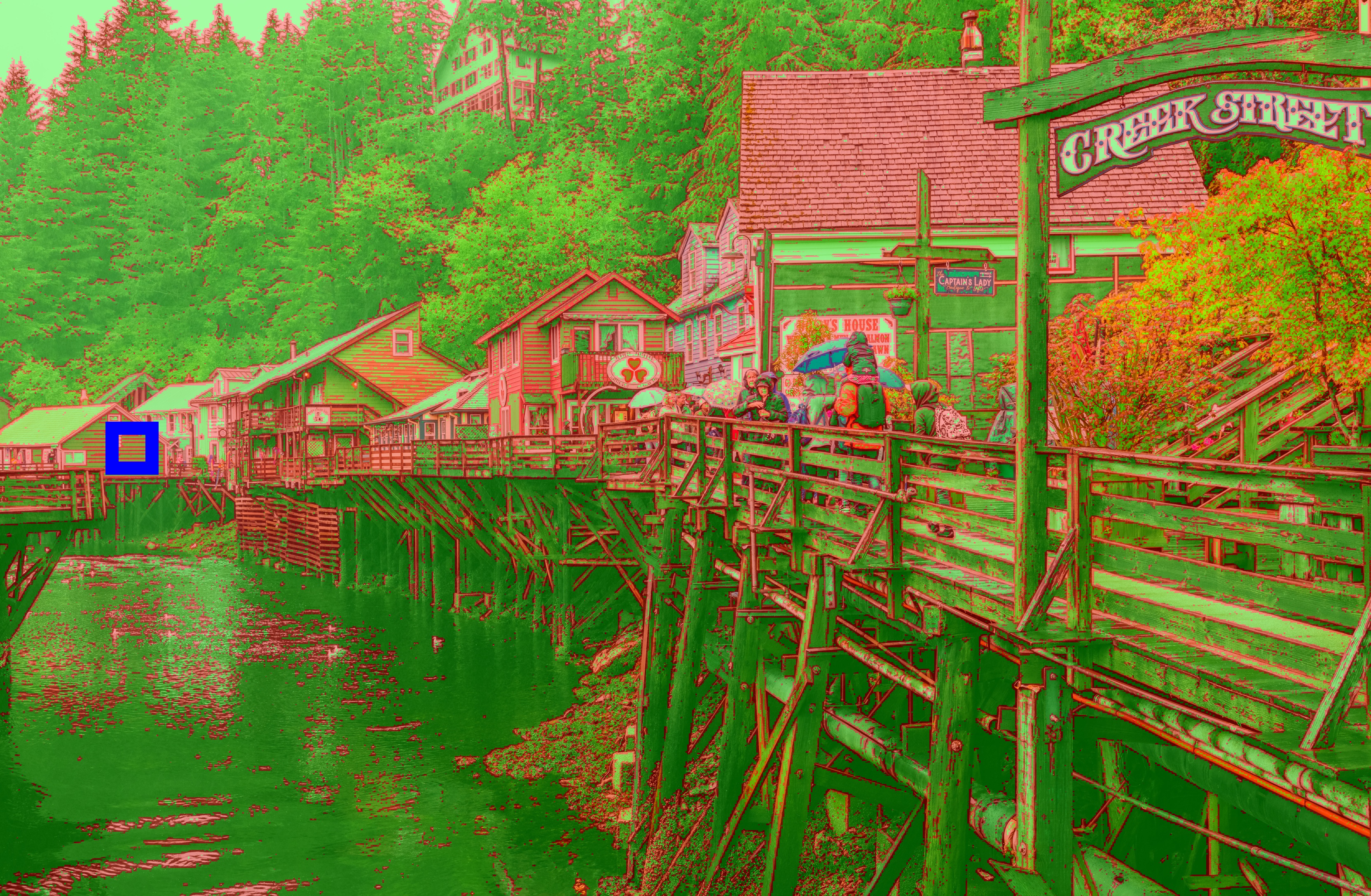}} &
\centering{25.07dB 416.2G(81\%)} &
\centering{25.04dB 451.1G(88\%)} &
\centering{25.16dB 322.2G(63\%)} &
\centering{25.20dB 511.2G(100\%)} &
\centering{} \tabularnewline
&
&
\includegraphics[width=\linewidth]{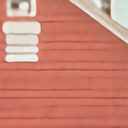} &
\includegraphics[width=\linewidth]{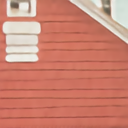} &
\includegraphics[width=\linewidth]{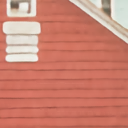} &
\includegraphics[width=\linewidth]{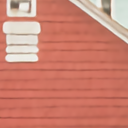} &
\includegraphics[width=\linewidth]{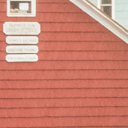} \tabularnewline
&
&
\includegraphics[width=\linewidth]{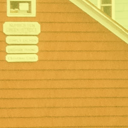} &
\includegraphics[width=\linewidth]{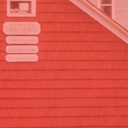} &
\includegraphics[width=\linewidth]{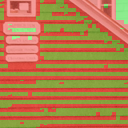} &
&
\tabularnewline
\midrule

\multirow{3}{*}[-6em]{\begin{subfigure}[m]{\linewidth}\caption{}\label{fig:visual_comparisons_4k_c}\end{subfigure}} &
\multirow{3}{*}[-4em]{\includegraphics[width=\linewidth]{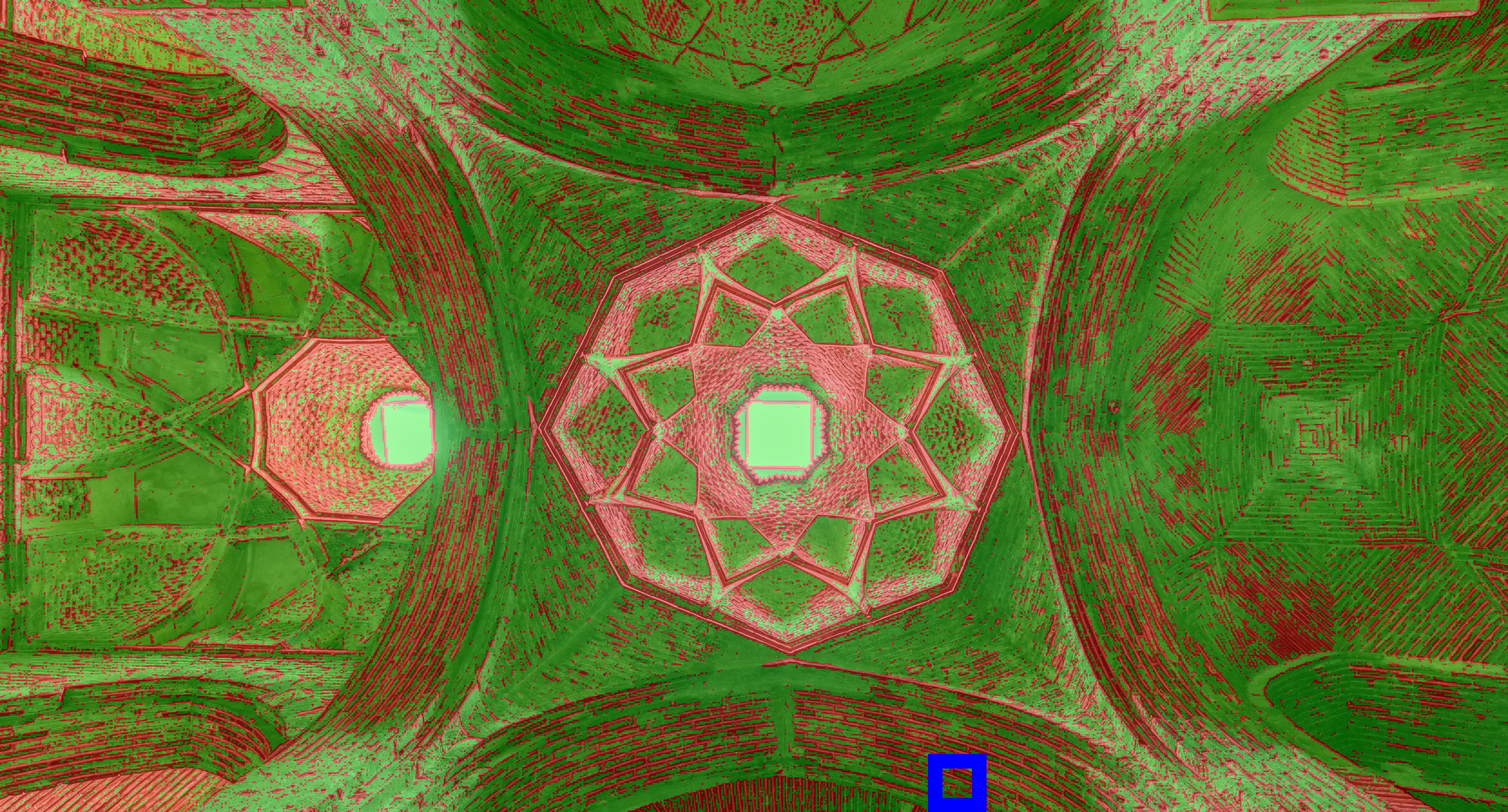}} &
\centering{27.08dB 317.7G(75\%)} &
\centering{27.10dB 366.8G(87\%)} &
\centering{27.15dB 265.5G(63\%)} &
\centering{27.20dB 421.0G(100\%)} &
\centering{} \tabularnewline
&
&
\includegraphics[width=\linewidth]{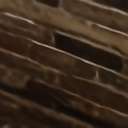} &
\includegraphics[width=\linewidth]{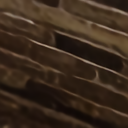} &
\includegraphics[width=\linewidth]{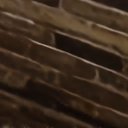} &
\includegraphics[width=\linewidth]{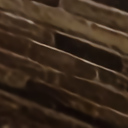} &
\includegraphics[width=\linewidth]{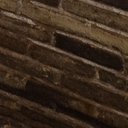} \tabularnewline
&
&
\includegraphics[width=\linewidth]{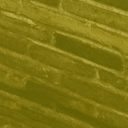} &
\includegraphics[width=\linewidth]{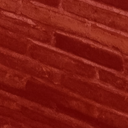} &
\includegraphics[width=\linewidth]{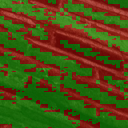} &
&
\tabularnewline
\midrule

\multirow{3}{*}[-6em]{\begin{subfigure}[m]{\linewidth}\caption{}\label{fig:visual_comparisons_4k_d}\end{subfigure}} &
\multirow{3}{*}[-4em]{\includegraphics[width=\linewidth]{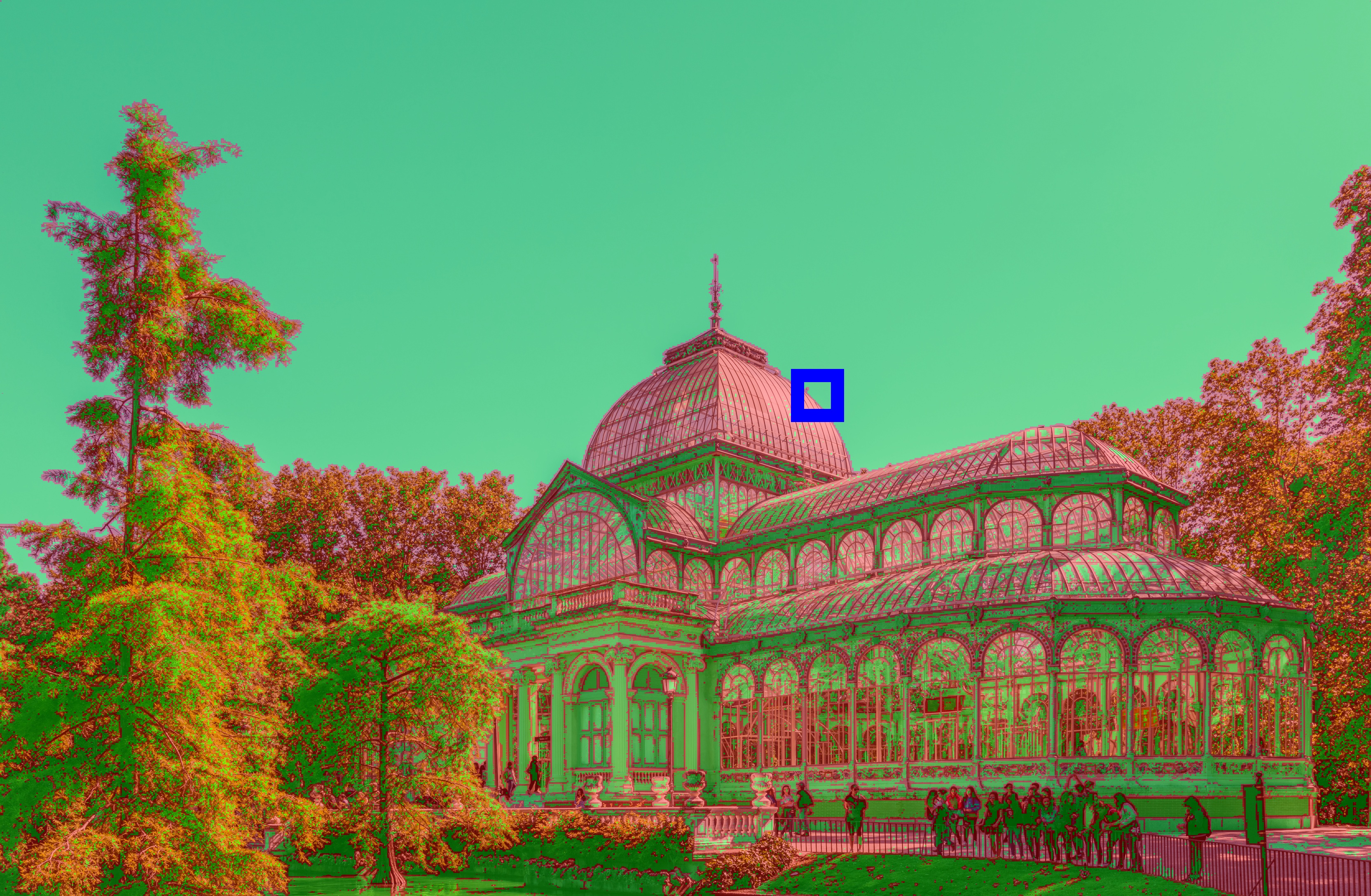}} &
\centering{24.74dB 351.0G(69\%)} &
\centering{24.71dB 291.1G(57\%)} &
\centering{24.76dB 311.4G(61\%)} &
\centering{24.81dB 511.2G(100\%)} &
\centering{} \tabularnewline
&
&
\includegraphics[width=\linewidth]{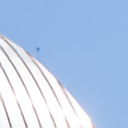} &
\includegraphics[width=\linewidth]{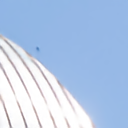} &
\includegraphics[width=\linewidth]{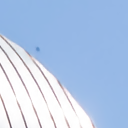} &
\includegraphics[width=\linewidth]{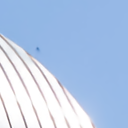} &
\includegraphics[width=\linewidth]{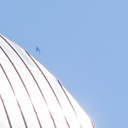} \tabularnewline
&
&
\includegraphics[width=\linewidth]{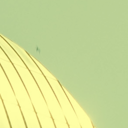} &
\includegraphics[width=\linewidth]{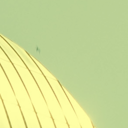} &
\includegraphics[width=\linewidth]{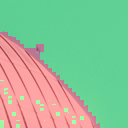} &
&
\tabularnewline

\end{tabular}
\caption{
Qualitative results of the previous methods \cite{chen2022arm,kong2021classsr} and our method with $\times$4 SR on Test4K.
}
\label{fig:visual_comparisons_test4k}
\end{figure}

In this section, we provide additional visual comparisons to ClassSR and ARM, along with PSNR values and FLOPs, demonstrating our method's efficiency and capability.
In Fig. \ref{fig:visual_comparisons_2k_b}, the patch-based methods engage in over-computation, which results in unnecessary computational expense. Our method saves computations by efficiently allocating resources on a pixel basis while maintaining high quality.
In Fig. \ref{fig:visual_comparisons_2k_c}, while under-computation by patch-based methods results in blurry outcomes, our method differentiates difficulties with precision, producing sharper and more defined restorations.
For Fig. \ref{fig:visual_comparisons_2k_d} and \ref{fig:visual_comparisons_4k_d}, instead of applying moderate computation uniformly across patches, our method focuses on challenging areas, achieving higher image quality with comparable computational cost.
Across various cases, the patch-based methods struggle with mixed restoration difficulties within a patch, but our pixel-level classification manages these variations effectively, improving both PSNR and FLOPs efficiency.
\begin{table}[H]
\centering
\caption{
Comparison of BSRN with and without PCSR on scale $\times$4 SR.
}
\label{tab:bsrn}
\resizebox{\linewidth}{!}{%
\begin{tabular}{|c|c|cc|cc|cc|}
\hline
Models             & Params. & Test2K(dB) & GFLOPs      & Test4K(dB) & GFLOPs      & Urban100(dB) & GFLOPs       \\ \hline
BSRN             & 352K       & 26.16        & 66.0 (100\%)  & 27.52      & 270.0 (100\%) & 24.43      & 29.1 (100\%)  \\
BSRN-PCSR         & 198K       & 26.10        & 51.8 (78\%)   & 27.52      & 208.4 (77\%)   & 24.29      & 23.6 (81\%)   \\ \hline
\end{tabular}
}
\end{table}

\end{document}